\documentclass[sigconf]{acmart}
\renewcommand\footnotetextcopyrightpermission[1]{} 

\AtBeginDocument{%
  \providecommand\BibTeX{{%
    \normalfont B\kern-0.5em{\scshape i\kern-0.25em b}\kern-0.8em\TeX}}}

\setcopyright{none}
\copyrightyear{}
\acmYear{}
\acmDOI{}
\acmConference[Document Intelligence Workshop at KDD]{The Second Document Intelligence Workshop at KDD}{2021}{Virtual Event}
\acmPrice{}
\acmISBN{}

\usepackage{graphicx}
\usepackage{hyperref}
\usepackage{booktabs}
\usepackage{multirow}
\usepackage{subcaption}
\usepackage{gensymb}
\usepackage{amsmath}
\usepackage{enumitem}
\usepackage{float}

\begin{document}

\title{Lights, Camera, Action! \\ A Framework to Improve NLP Accuracy over OCR documents}


\author{Amit Gupte}\authornote{These authors contributed equally to this research.}
\author{Alexey Romanov}\authornotemark[1]
\author{Sahitya Mantravadi}\authornotemark[1]\email{{amgupte,samantr,alromano}@microsoft.com}
\affiliation{%
  \institution{MAIDAP, Microsoft}
  \city{Cambridge}
  \state{MA}
  \country{USA}
}

\author{Dalitso Banda}\authornotemark[1]
\author{Jianjie Liu}\authornotemark[1]\email{{jianjl,dabanda}@microsoft.com}
\affiliation{%
  \institution{MAIDAP, Microsoft}
  \city{Cambridge}
  \state{MA}
  \country{USA}
}

\author{Raza Khan}\authornote{Currently working at Amazon Alexa.}
\email{razaurrehman@gmail.com}
\affiliation{%
  \institution{Azure AI, Microsoft}
  \city{Bellevue}
  \state{WA}
  \country{USA}
}

\author{Lakshmanan Ramu Meenal}\email{laramume@microsoft.com}
\affiliation{%
  \institution{Azure AI, Microsoft}
  \city{Bellevue}
  \state{WA}
  \country{USA}
}
\author{Benjamin Han}\email{diha@microsoft.com}
\affiliation{%
  \institution{Azure AI, Microsoft}
  \city{Bellevue}
  \state{WA}
  \country{USA}
}

\author{Soundar Srinivasan}\email{sosrini@microsoft.com}
\affiliation{%
  \institution{MAIDAP, Microsoft}
  \city{Cambridge}
  \state{MA}
  \country{USA}
}

\renewcommand{\shortauthors}{Gupte, Romanov, Mantravadi, Banda, Liu, Khan, Meenal, Han \& Srinvasan}

\definecolor{BrickRed}{rgb}{0.8, 0.25, 0.33}
\definecolor{ForestGreen}{rgb}{0.0, 0.27, 0.13}
\definecolor{RoyalBlue}{rgb}{0.25, 0.41, 0.88}

\begin{abstract}
   Document digitization is essential for the digital transformation of our societies, yet a crucial step in the process, Optical Character Recognition (OCR), is still not perfect. Even commercial OCR systems can produce questionable output depending on the fidelity of the scanned documents. In this paper, we demonstrate an effective framework for mitigating OCR errors for any downstream NLP task, using Named Entity Recognition (NER) as an example. We first address the data scarcity problem for model training by constructing a document synthesis pipeline, generating realistic but degraded data with NER labels. 
   We measure the NER accuracy drop at various degradation levels and show that a text restoration model, trained on the degraded data, significantly closes the NER accuracy gaps caused by OCR errors, including on an out-of-domain dataset. For the benefit of the community, we have made the document synthesis pipeline available as an open-source project.
   
\end{abstract}




\maketitle

\section{Introduction}
\label{sec:intro}
Despite years of advancement of information technologies, a vast amount of information is still locked inside \textit{analog} documents. For example, the JFK Assassination Records Collection\footnote{\url{https://www.archives.gov/research/jfk}} consists of over 5 million pages of records. Document digitization technologies like Optical Character Recognition (OCR) have made it possible to index the collection at a \textit{word} level, allowing search via keywords and phrases. More elaborate NLP technologies such as Named Entity Recognition (NER) can then be applied to extract information such as person names, allowing information retrieval at \textit{entity} level.
However, even modern OCR technologies can produce output of a poor quality depending on the quality of the scanned documents.

In this paper, we demonstrate an effective framework for mitigating the impact of OCR errors on any downstream NLP task, using the task of NER as an example. Although there is a growing body of work dedicated to OCR and quality improvements, the specific topic of the impact of OCR errors on NER has not been widely explored. With the exception of ~\cite{hamdi2019analysis,hwang2019postocr}, and to an extent~\cite{jean2017lexicographical}, the majority of the previous work has focused on general OCR error detection
and correction~\cite{hakala2019leveraging,d2017generating}. Here, we focus on a framework that allows us to improve the accuracy of any downstream NLP task such as NER.

Our major contributions are as follows. (1) To address the scarcity of entity-labeled OCR documents for model training, we design a pipeline \texttt{Genalog} for \underline{gen}erating an\underline{alog} synthetic documents. The pipeline takes plain texts optionally annotated with named entities, synthesizes degraded document images, runs OCR on the images, and propagates the labels onto the OCR output texts. These imperfect texts can then be aligned with their clean counterparts for model training. (2) We propose an \textit{action prediction} model that can restore clean text from OCR output and mitigate the downstream NER accuracy degradation. (3) We systematically investigate NER accuracy drop on OCR output at various synthetic degradation levels, and show that our text restoration model can indeed significantly close the NER accuracy gap.

The core capabilities of \texttt{Genalog} -- synthesizing visual document images, degrading images, extracting text from images, and performing text alignment with label propagation -- can have a wide array of applications. To facilitate further research, we are making \texttt{Genalog} available as an open-source project on GitHub\footnote{\url{https://github.com/microsoft/genalog}}.


\section{Related Work}
\label{sec:related}


Since we focus on closing the accuracy gap of NER induced by OCR errors, we adopt a more conventional NER model and keep it constant in our experimentation, instead of using the most recent transformer-based ~\cite{vaswani2017attention} NER model such as~\cite{wang2019cross}.
LSTM-CNN models ~\cite{lstmcnnner} typically employ a token-level Bi-LSTM on top of a character-level CNN layer. Hence, we train a multi-task Bi-LSTM model that operates at both levels ~\cite{wang2019cross}. This provides more granularity at the character-level than the traditional Bi-LSTM CRF model~\cite{huang2015bidirectional}. 


To systematically study the challenges of performing NER on OCR output, \cite{hamdi2019analysis} categorized four types of OCR degradations and examined the decrease in NER accuracy for each as evaluated on synthetic documents. Their analysis referenced DocCreator \cite{journet2017doccreator} as a tool for generating synthetic documents. Our implementation, \texttt{Genalog}, was initially inspired by this tool. Another work, ~\cite{syntheticrecipeocr}, presented a flexible framework for synthetic document generation but did not produce annotations for a downstream task. 

The work of \cite{strien2020ocrimpact} examined the impact of noise introduced by OCR on analog documents on several downstream NLP tasks, including NER. They observed a consistent relationship between decreased OCR quality and worse NER accuracy. ~\cite{miller2000named} explored the relationship between the word error rates of noisy speech and OCR on downstream NER, and ~\cite{packer2010extracting} noted the difficulty of extracting names from noisy OCR text.


The goal of our text restoration approach is to reconstruct a ``clean'' version of the text from OCR output, which may contain various misspellings and errors. Sequence-to-sequence (seq2seq) models
are a natural first approach: they convert an input sequence into an output sequence and can process text at the character level~\cite{hakala2019leveraging}. The use of LSTMs~\cite{hochreiter1997long} is a common paradigm for seq2seq models.
Another approach is the use of an encoder-decoder model with a single or multiple LSTM layers~\cite{d2017generating,suissa2020optimizing}. Here, the encoder-decoder model is not based on a seq2seq architecture; instead it is a straightforward decoding of a character at every time step corresponding to the input sequence.
These approaches are powerful but have a few drawbacks that we will explore in details in Section~\ref{sec:restoration_model}.

\section{Generation of Synthetic Documents}
\label{sec:genalog}

\begin{figure}[htp]
    \centering
    \includegraphics[clip, trim=3cm 9cm 5cm 4cm, width=0.95\linewidth]{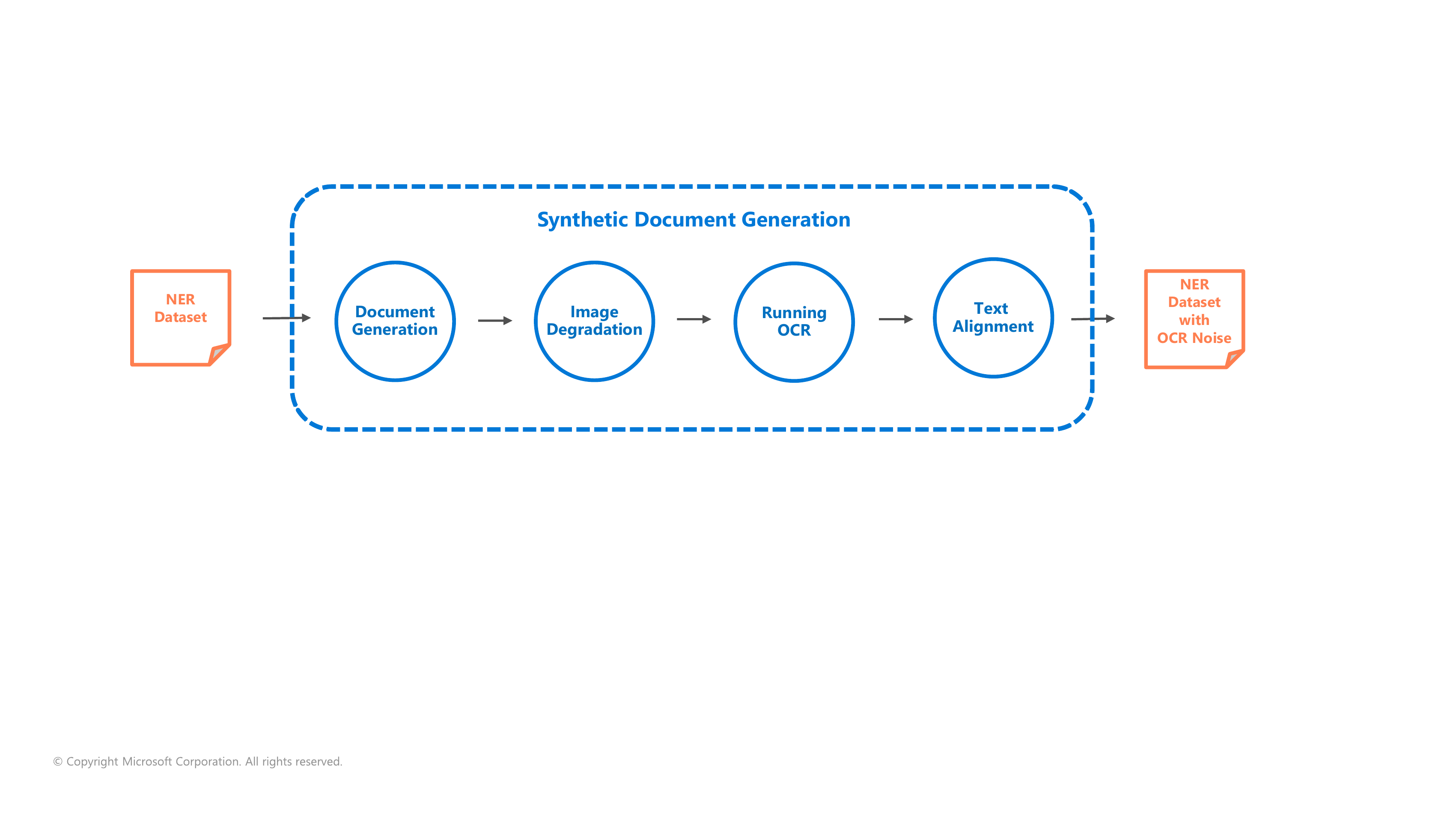}
    \caption{Synthetic document generation pipeline}
    \label{fig:synthetic_doc_gen}
\end{figure}

While there are many publicly available annotated datasets for NER, there are few targeting OCR output, i.e., document images containing texts with entities marked manually. To address this data scarcity, we developed \texttt{Genalog}, a Python package to synthesize document images from text with customizable degradation. It can also run an OCR engine on the images and align the output with ground truth to propagate the NER labels. The final result consists of degraded document images, corresponding OCR text, and NER labels for the OCR text suitable for model training and testing  (\autoref{fig:synthetic_doc_gen}). We now describe each of these steps in the remainder of this section.

\subsection{Document Generation}
A document contains various layout information such as font family/size, page layout, etc. Our goal is to generate a document image given a specified layout and input text. \texttt{Genalog} provides several standard document templates implemented in HTML/CSS (shown in~\autoref{sec:document-generation}), and a browser engine is used to render document images.The layout can be reconfigured via CSS properties, and a template can be extended to include other content such as images and tables. In our experiments, we use a simple text-block template.
  
\subsection{Image Degradation}
\texttt{Genalog} supports elementary degradation operations such as Gaussian blur, bleed-through, salt and pepper, and morphological operations including open, close, dilate, and erode. 
Each effect can be applied at various strengths, and multiple effects can be stacked together to simulate realistic degradation. \autoref{fig:appendix-degradation-examples}
provides more details on each degradation.

\subsection{Running OCR}
We use a commercial OCR API
to extract text from document images. \texttt{Genalog} calls the service on batches of documents and obtains extracted lines of text and their bounding boxes on each page. \texttt{Genalog} also computes metrics measuring the OCR accuracy, including Character Error Rate (CER), Word Error Rate (WER), and two additional classes of metrics: edit distance~\cite{levenshtein} and alignment gap metrics. 
This provides information on OCR errors and can capture errors such as \textbf{``room"} misrecognized as \textbf{``roorn"} by OCR.

\subsection{Text Alignment}
\label{sec:text_alignment}


After obtaining the OCR text from synthetic documents, we propagate NER labels from the source text to the OCR text by aligning these texts and propagating the labels accordingly. \texttt{Genalog} uses Needleman-Wunsch algorithm~\cite{needleman1970general} for text alignment on the character level. Because the algorithm is $O(n^2)$ in both time and space complexity, it can be inefficient on long documents. To improve efficiency, we use the Recursive Text Alignment Scheme~\cite{retas} and search for unique words common to both ground truth and OCR text as anchor points to break documents into smaller fragments for faster alignment, thereby 
obtaining a 15-20x speed-up on documents with average length of 4000 characters.

\subsection{Degradation Effect on OCR Error Rates}
To understand what degradation on synthetic documents is realistic, we compute CER and WER on OCR texts obtained from a large corpus of \textit{real} document scans (first row of ~\autoref{tab:ocr_table}).
We then synthesize documents from the public CoNLL 2003~\cite{sang2003introduction} and CoNLL 2012~\cite{pradhan2012conll} datasets with different degrees of degradation (none, light and heavy) and compute CER and WER on the OCR output. The result shows that in terms of WER, the \textit{light} degradation is closer to the real degradation than the \textit{heavy} degradation.

\begin{table}[htp]
    \centering
    \resizebox{0.85\linewidth}{!}{%
    \begin{tabular}{cccc} \toprule
    Dataset & Degradations & CER (\%) & WER(\%)\\\midrule
    Real    & - & 1.2 &   7.3 \\
    CoNLL '03 & None & 0.3 & 2.5 \\
    CoNLL '12 & None & 0.3 & 2.5 \\
    CoNLL '03 & All (light) & 0.5 & 5.4 \\
    CoNLL '12 & All (light) & 0.9 & 7.5 \\
    CoNLL '03 & All (heavy) & 9.7 & 36.6 \\
    CoNLL '12 & All (heavy) & 8.5 & 33.4 \\
    \bottomrule
    \end{tabular}
    }
    \caption{Error rates on Real and Synthetic OCR data}
    \label{tab:ocr_table}
\end{table}

\begin{figure}[htp]
    \centering
    \subcaptionbox{Bottom: OCR output. Top: ground truth. 
    Notice the growing character shift.
    \label{fig:charcter_shift_problem}}[0.48\linewidth]{
        \includegraphics[clip, trim=0.1cm 0.3cm 0.1cm 0.3cm, width=0.95\linewidth]{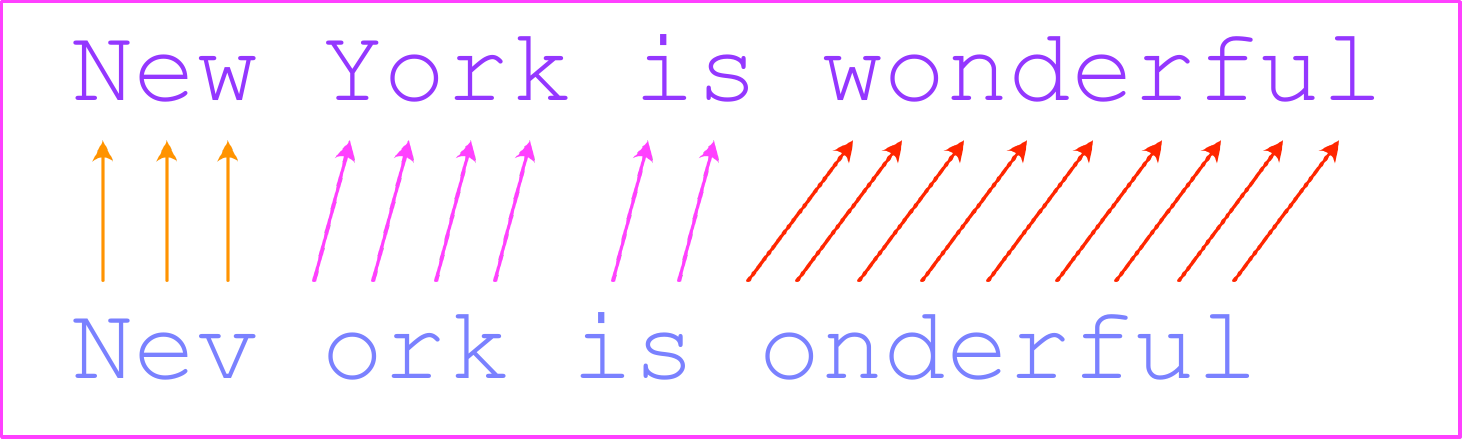}
    }
    \hfill
    \subcaptionbox{Bottom: OCR output. Top: target characters to restore.
    Action prediction does not have character shift.
    \label{fig:charcter_shift_actions}}[0.48\linewidth]{
        \includegraphics[clip, trim=0.1cm 0.3cm 1.5cm 0.3cm, width=0.95\linewidth]{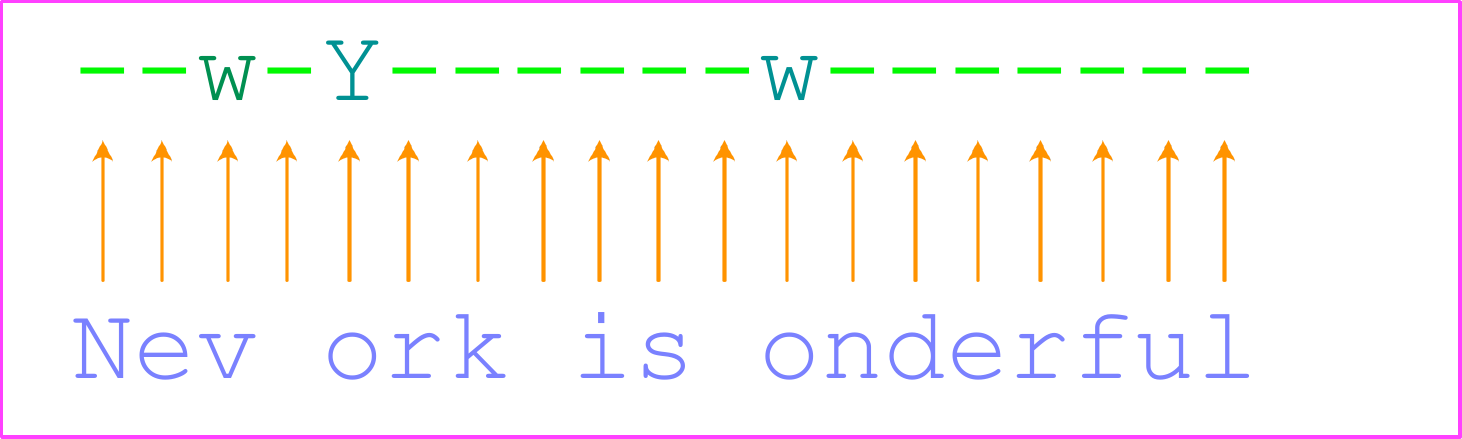}
    }
    \caption{Problem of character shift and its mitigation.}
    \label{fig:charcter_shift}
\end{figure}

\section{Text Restoration Model}
\label{sec:restoration_model}
To make our framework for mitigating OCR errors flexible, we propose a model for \textit{text restoration}. This model can be trained independently from the downstream task, and training data can be readily obtained from a data synthesis pipeline such as \texttt{Genalog}.

One approach to generate a corrected sequence of text is to correct one word at a time via a seq2seq model such as ~\cite{hakala2019leveraging}. In that work they report when allowing an entire sentence as input instead of a single word, the model trained with smaller dataset resulted in \textit{much increased} WER.

An alternative approach is to adopt an encoder-decoder model that decodes one character a time in sync with the input. Although this approach is faster than seq2seq, as insertions or deletions accumulate in a long sequence, the model needs to account for a growing character shift (see~\autoref{fig:charcter_shift_problem}). This issue often leads to a repetition of the same character for several timesteps during the prediction. \cite{d2017generating} mitigated this problem by limiting the length of the sequence to 20 characters and utilizing a sliding window. This significantly slows down inference and limits the context of the model.

\begin{figure}
    \begin{subfigure}[b]{.48\linewidth}
        \centering
        \includegraphics[clip, trim=0cm 0cm 0cm 0cm, width=0.95\linewidth]{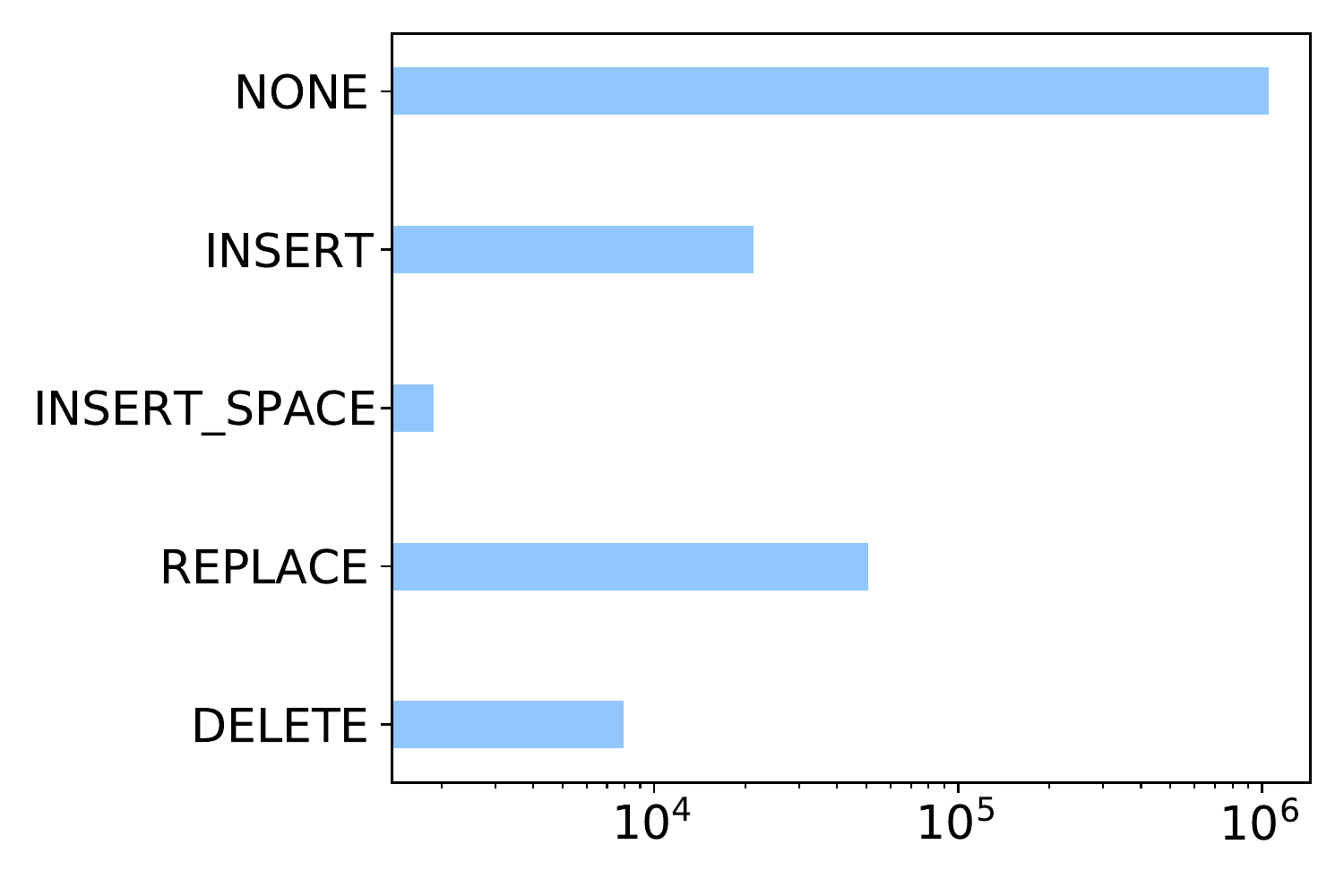}
        \caption{Distribution of actions}
        \label{fig:actions_dist}
    \end{subfigure}%
    \begin{subfigure}[b]{.48\linewidth}
        \centering
        \includegraphics[clip, trim=0cm 0cm 0cm 0cm, width=0.95\linewidth]{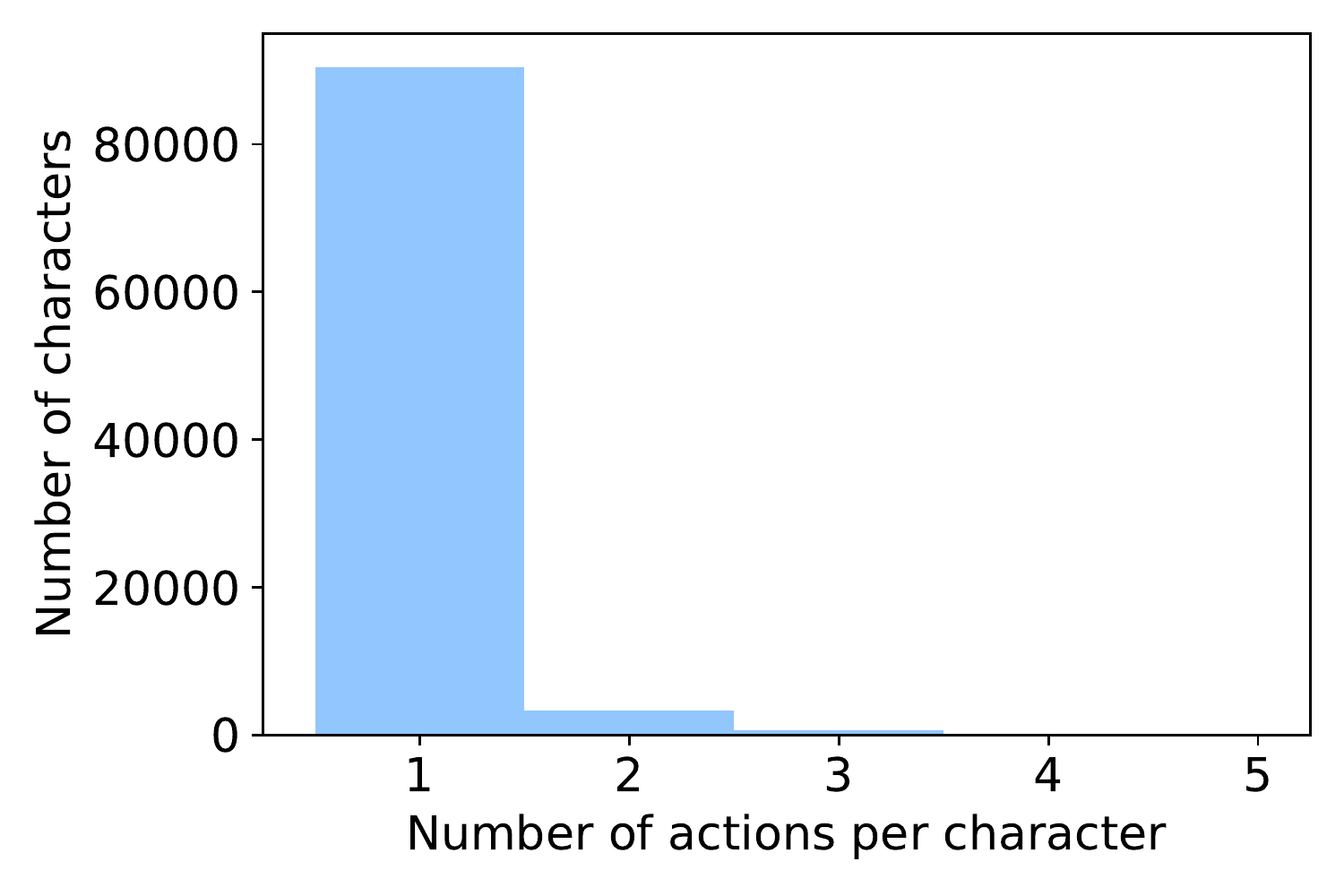}
        \caption{Actions per character}
        \label{fig:characters_actions_dist}

    \end{subfigure}
    \caption{Actions on CoNLL-2012, All Degradations}
    \label{fig:actions_dist_charts}
\end{figure}

Our solution builds on the previous approaches, but instead of predicting characters at each time step, we predict \textit{actions} that are required for restoration (\autoref{fig:charcter_shift_actions}). For example, to restore the source text ``Cute cat'' from the OCR output ``Cute at'', we need to \textit{insert} a ``c'' before the ``a''. This is akin to sequence labelling: each character in the input sentence is assigned an action labe). There are four main actions (\texttt{INSERT}, \texttt{REPLACE}, \texttt{INSERT\_SPACE}, and \texttt{DELETE}) and two auxiliary actions (\texttt{NONE} and \texttt{PAD}). The first three actions need a character, e.g., we need to specify which character to \texttt{INSERT}. Predicting actions and characters separately can reduce vocabulary size and label sparsity, which can be severe for languages such as Chinese or Japanese. 
The distributions of the actions on CoNLL-2012 is presented in~\autoref{fig:actions_dist}. 

Since it is a sequence labelling problem, each input character is limited to have only one action. If a sentence is missing several characters in a row, our model is limited to recovering only one character. \cite{omelianchuk2020gector} and~\cite{awasthi2019parallel}, who suggested similar approaches for grammatical error correction, mitigate this by applying the model up to 3 times. 
In OCR we've observed that errors mostly occur in non-adjacent locations (see~\autoref{fig:characters_actions_dist}), and only few characters suffer from errors that need more than one action to fix.
This is also supported by~\cite{jatowt2019deep}.
After a manual examination of the characters that need exactly two actions, we found that most of them need the following two \texttt{INSERT} actions: insert a space and a character. We therefore introduce an action \texttt{INSERT\_SPACE} that inserts both in one action.

\begin{figure}[ht]
    \centering
    \includegraphics[clip, trim=0.1cm 0.1cm 0.1cm 0.1cm, width=0.7\linewidth]{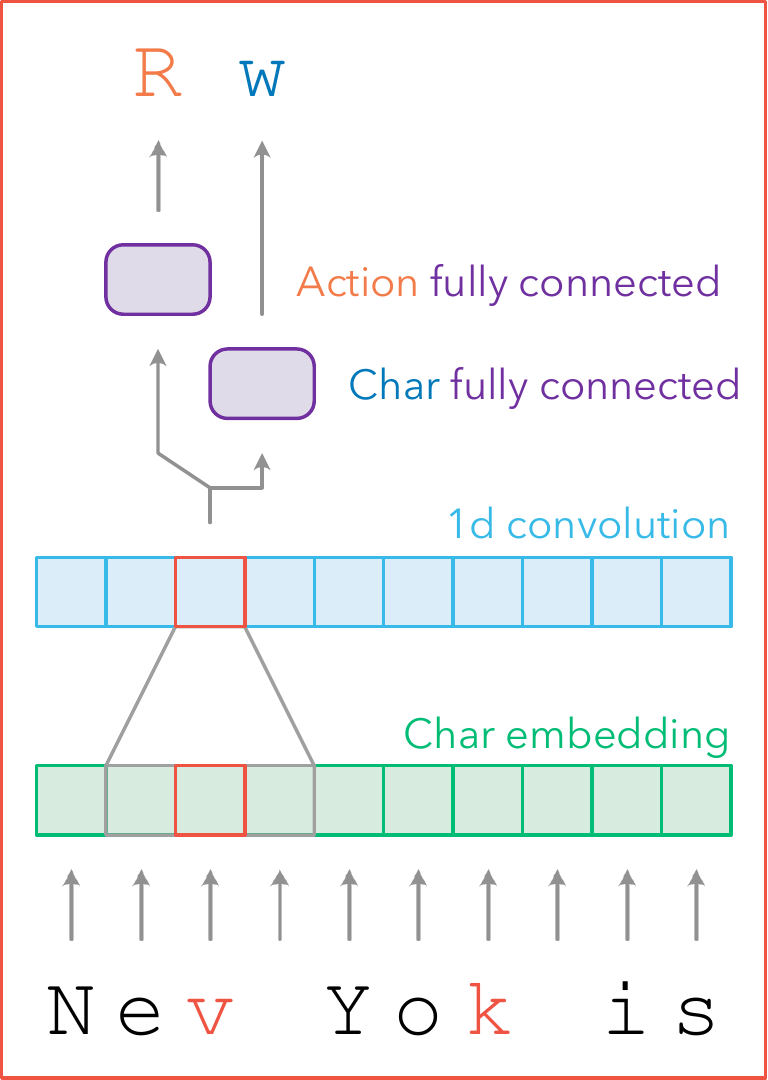}
    \caption{Action Prediction Model Architecture}
    \label{fig:model_architecture}
\end{figure}

Since OCR errors usually have single erroneous characters, fixing them does not require long contexts. Our model thus consists of a character embedding layer followed by one-dimensional convolution layers, and for each input character, it predicts an action and a character using two separate fully-connected layers. The architecture of the model is presented on~\autoref{fig:model_architecture}.
This model is trained with a weighted combination of cross-entropy losses for actions and characters:
$\mathcal{L}_\text{total} = \alpha * \mathcal{L}_a + \beta * \mathcal{L}_c$.

\section{Experiments and Results}
\label{sec:expts}

\begin{table*}[htp]
\centering
\resizebox{\textwidth}{!}{%
\begin{tabular}{@{}lll@{}}
\toprule
OCR Sentence & Reconstructed Sentence & Ground Truth Sentence \\ \midrule
No one will be able to \textcolor{BrickRed}{rec- ognize} her body ! & No one will be able to \textcolor{ForestGreen}{recognize} her body . & No one will be able to \textcolor{RoyalBlue}{recognize} her body ! \\
Although \textcolor{BrickRed}{!} still have n't made the final decision to go & Although \textcolor{ForestGreen}{I} still have n't made the final decision to go & Although \textcolor{RoyalBlue}{I} still have n't made the final decision to go \\
Why do you think the North Koreans chose \textcolor{BrickRed}{july Fourth} /? & Why do you think the North Koreans chose \textcolor{ForestGreen}{July Fourth} /? & Why do you think the North Koreans chose \textcolor{RoyalBlue}{July Fourth} /? \\
\textcolor{BrickRed}{israel} says \textcolor{BrickRed}{\&} may abandon the peace negotiations altogether . & \textcolor{ForestGreen}{Israel} says \textcolor{ForestGreen}{it} may abandon the peace negotiations altogether . & \textcolor{RoyalBlue}{Israel} says \textcolor{RoyalBlue}{it} may abandon the peace negotiations altogether . \\
\textcolor{BrickRed}{!} will protect this \textcolor{BrickRed}{cky} and save it . & \textcolor{ForestGreen}{I} will protect this \textcolor{ForestGreen}{city} and save it . & \textcolor{RoyalBlue}{I} will protect this \textcolor{RoyalBlue}{city} and save it . \\ \bottomrule
\end{tabular}%
}
\caption{Sample of sentences that were correctly restored by the action prediction model}
\label{tbl:restoration examples}
\end{table*}

To investigate the effect of OCR errors on NER accuracy and the effectiveness of our model in mitigating them, we first use \texttt{Genalog} to create a synthetic dataset from clean texts (``clean data'') and propagate the NER annotations to the OCR text (``noisy data''). We measure the the accuracy of the NER model on the clean data and the noisy data. 
Using the alignment between the ground truth and the OCR output, we produce labeled data to train the text restoration model. We then evaluate the same NER model on the restored text to determine if we can close the accuracy gap caused by OCR.



\subsection{Data}
We use the well-known corpora from CoNLL-2003~\cite{sang2003introduction} and CoNLL-2012~\cite{pradhan2012conll} and refer to them as ``clean data". Next, \texttt{Genalog} synthesizes and generates equivalent OCR texts, called ``noisy data". For each degradation effect
in \texttt{Genalog}, we generate three versions of noisy data: none, light, or heavy degradation. This allow us to observe, at a finer-grained level, the influence of degradations on NER accuracy. 
We also produce a version of datasets with \textit{All Degradations}. ~\autoref{sec:image-degradation} shows example images of each degradations and \autoref{sec:appendix-degradation-params} lists degradation  parameters.

\subsection{NER Model}
We use a popular character-level and token-level Bi-LSTM model~\cite{dernoncourt2017neuroner} for NER. The model consists of a char-level Bi-LSTM layer followed by a token-level Bi-LSTM. Since we are using CoNLL-2012 and CoNLL-2003 to train the model, we utilize separate classification layer for each dataset~\cite{wang2019cross}. 
We settled on this popular approach since our focus is mitigating NER accuracy drop caused by OCR, not achieving state-of-the-art NER on clean text.

\subsection{Results and Analysis}
\label{sec:results}
In this section, we report the evaluation results of our action prediction model, analyze the effect of different degradations on NER accuracy, and show how our model behaves on out-of-domain datasets.

\paragraph{Accuracy of the Action Prediction Model}
\autoref{tbl:restoation_model_acc} compares character-level and word-level accuracy between the noisy OCR text and the restored text. We observe that our model improves accuracy according to both metrics, and small improvements at character level lead to bigger improvements at word level.


\begin{table}[ht]
\centering
\resizebox{0.9\linewidth}{!}{%
\begin{tabular}{@{}lllll@{}}
\toprule
Dataset & \multicolumn{2}{l}{CoNLL-2012} & \multicolumn{2}{l}{CoNLL-2003} \\
Degradations & Light & Heavy & Light & Heavy \\ \midrule
Char Accuracy Noisy & 0.986 & 0.900 & 0.989 & 0.900 \\
Char Accuracy Restored & 0.991 & 0.907 & 0.994 & 0.903 \\
Word Accuracy Noisy & 0.927 & 0.661 & 0.942 & 0.646 \\
Word Accuracy Restored & 0.962 & 0.732 & 0.969 & 0.700 \\ \bottomrule
\end{tabular}
}
\caption{Character-level and word-level accuracy of the noisy text and restored text}
\label{tbl:restoation_model_acc}
\end{table}

There are several common OCR errors that our model is able to recover, including errors introduced by line breaks and hyphenation (e.g., ``rec-ognize'' to ``recognize''). Remarkably, it also restores words suffering from multi-character errors, despite the one action per character limitation.
\autoref{tbl:restoration examples} shows more examples.

\paragraph{NER Accuracy on Noisy Text}
To improve NER accuracy on noisy text, we first pass it through the action prediction model trained on both CoNLL-2003 and CoNLL-2012. The restored text is then fed to the NER model to generate entities. To measure the benefit in NER accuracy due to the use of restoration, we first examine the drop in NER accuracy introduced by degradation and OCR.
~We then evaluate NER on the restored text. By comparing the NER accuracy on the noisy text to that on the restored text, we can then compute the improvement introduced by the restoration model. \autoref{tbl:results_light_without_cnn} shows the result of this evaluation using All Degradations (Light) on CoNLL-2012 and CoNLL-2003. The NER accuracy on noisy text goes down significantly on all datasets. Our restoration approach is able to improve the F1 scores considerably, and the accuracy gaps are reduced up to 73\% on CoNLL datasets. 

\begin{table}[htp]
\centering
\resizebox{0.9\linewidth}{!}{%
\begin{tabular}{@{}llll@{}}
\toprule
Dataset & CoNLL-2012 & CoNLL-2003 & CNN \\ \midrule
Clean Text                              & 0.832      & 0.860      & 0.989 \\
Noisy Text                           & 0.783      & 0.820      & 0.590 \\
Restored Text                           & 0.819      & 0.841      & 0.895 \\
Relative Gap Reduction                  & 73\%       & 52\%       & 76\%  \\ \bottomrule
\end{tabular}%
}
\caption{NER F1 score on the clean, noisy, and restored text. See \autoref{sec:results} for the description of the CNN column.
} 
\label{tbl:results_light_without_cnn}
\end{table}

\begin{table}[htp]
\centering
\resizebox{\linewidth}{!}{%


\begin{tabular}{@{}llllllll@{}}
\toprule
Degradation     & None      & Bleed & Blur  & Pepper    & Phantom   & Salt  & All \\ \midrule
Noisy Text      & 0.807     & 0.561 & 0.780 & 0.742     & 0.795     & 0.749 & 0.517 \\
Restored Text   & 0.828     & 0.656 & 0.813 & 0.785     & 0.824     & 0.798 & 0.576 \\
Gap Reduction   & 84\%      & 35\%  & 64\%  & 47\%      & 77\%      & 59\%  & 19\%  \\ \bottomrule
\end{tabular}%


}
\caption{NER F1 score on the noisy and the restored text from the CoNLL-2012 test set with heavy degradation. 
}
\label{tbl:results_ner_horizontal}
\end{table}

To further investigate the effect of \textit{heavy} degradations, we applied these degrations to CoNLL-2012 dataset and measured the resulting errors in NER. \autoref{tbl:results_ner_horizontal} shows that all degradations hurt the NER accuracy and, in all cases, the proposed restoration model helps to improve the NER scores. The NER accuracy on degraded text declines the most when all degradations are combined, but
19\% of this gap is recovered by our model. In general, we notice that the action prediction model works the best at moderate levels of degradation, where up to 77\% of the accuracy drop is restored. When the text is obscured by many degradations in the heavy mode, the gap reduction becomes smaller.

\paragraph{NER Accuracy on Out-of-domain Datasets}
To evaluate how our model generalizes to unseen documents from another domain, we evaluated our framework on a new\footnote{Although CoNLL includes news articles, they are just a part of it.} ``CNN'' dataset -- 1000 randomly sampled articles from the CNN part of the DeepMind Q\&A Dataset~\cite{hermann2015teaching}. 
Since the CNN datset does not have NER labels, we run our NER model on it to produce the ``ground truth''. We then use \texttt{Genalog} to generate synthetic documents with \texttt{All Degradations (Light)}.

The accuracy of the NER system on the degraded text drops from the perfect score (the ground truth labels are produced by the same system)\footnote{It is less than 1.0 due to sentence splitting differences between label generation and evaluation.} to the F1 score of 0.59. Notice that the NER accuracy gap on the CNN dataset is much bigger than that on the CoNLL datasets (40 vs. 4.5 points), because here we are evaluating the accuracy using the model output on clean text as ground truth. Any change in NER prediction due to OCR errors therefore will contribute to the accuracy gap, while in the case of CoNLL datasets, changing an \textit{incorrect} NER prediction (based on human annotations) to another incorrect prediction does not contribute to the accuracy gap.

After applying the restoration model to the degraded text, the accuracy of the NER system rises to 0.895 F1. Despite not being trained on the CNN dataset, our action prediction model is still able to close 76\% of the NER accuracy gap. This result shows the action prediction model is able to generalize to data from another domain.

\section{Conclusion}
\label{sec:conclusion}

In this paper, we have demonstrated an effective framework to mitigate errors produced by OCR, a key step in the process of document digitization. We first constructed a data synthesis pipeline, \texttt{Genalog}, capable of generating synthetic document images given plain text input, running the images through OCR, and propagating annotated NER labels from the input text to the OCR output. We then proposed a novel approach, the \textit{action prediction model}, to restore text from the inflicted OCR errors and show that our model does not suffer from problems faced by the conventional models when alignment mismatches between input and output accumulate. Lastly, we demonstrated that the accuracy of an important downstream task, NER, does drop at various degradation levels, and our text restoration model can significantly close the accuracy gaps, including on an out-of-domain dataset. Since the design of our restoration model does not depend on a the downstream task, it is generalizable to many other NLP tasks as well.


\bibliographystyle{ACM-Reference-Format}
\bibliography{main}


\begin{thebibliography}{26}


\ifx \showCODEN    \undefined \def \showCODEN     #1{\unskip}     \fi
\ifx \showDOI      \undefined \def \showDOI       #1{#1}\fi
\ifx \showISBNx    \undefined \def \showISBNx     #1{\unskip}     \fi
\ifx \showISBNxiii \undefined \def \showISBNxiii  #1{\unskip}     \fi
\ifx \showISSN     \undefined \def \showISSN      #1{\unskip}     \fi
\ifx \showLCCN     \undefined \def \showLCCN      #1{\unskip}     \fi
\ifx \shownote     \undefined \def \shownote      #1{#1}          \fi
\ifx \showarticletitle \undefined \def \showarticletitle #1{#1}   \fi
\ifx \showURL      \undefined \def \showURL       {\relax}        \fi
\providecommand\bibfield[2]{#2}
\providecommand\bibinfo[2]{#2}
\providecommand\natexlab[1]{#1}
\providecommand\showeprint[2][]{arXiv:#2}

\bibitem[\protect\citeauthoryear{Awasthi, Sarawagi, Goyal, Ghosh, and
  Piratla}{Awasthi et~al\mbox{.}}{2019}]%
        {awasthi2019parallel}
\bibfield{author}{\bibinfo{person}{Abhijeet Awasthi}, \bibinfo{person}{Sunita
  Sarawagi}, \bibinfo{person}{Rasna Goyal}, \bibinfo{person}{Sabyasachi Ghosh},
  {and} \bibinfo{person}{Vihari Piratla}.} \bibinfo{year}{2019}\natexlab{}.
\newblock \showarticletitle{Parallel Iterative Edit Models for Local Sequence
  Transduction}. In \bibinfo{booktitle}{\emph{Proceedings of the 2019
  Conference on Empirical Methods in Natural Language Processing and the 9th
  International Joint Conference on Natural Language Processing
  (EMNLP-IJCNLP)}}. \bibinfo{pages}{4251--4261}.
\newblock


\bibitem[\protect\citeauthoryear{Chiu and Nichols}{Chiu and Nichols}{2016}]%
        {lstmcnnner}
\bibfield{author}{\bibinfo{person}{Jason Chiu} {and} \bibinfo{person}{Eric
  Nichols}.} \bibinfo{year}{2016}\natexlab{}.
\newblock \showarticletitle{Named Entity Recognition with Bidirectional
  LSTM-CNNs}.
\newblock \bibinfo{journal}{\emph{Transactions of the Association for
  Computational Linguistics}} \bibinfo{volume}{4}, \bibinfo{number}{0}
  (\bibinfo{year}{2016}), \bibinfo{pages}{357--370}.
\newblock
\showISSN{2307-387X}
\urldef\tempurl%
\url{https://transacl.org/ojs/index.php/tacl/article/view/792}
\showURL{%
\tempurl}


\bibitem[\protect\citeauthoryear{Dernoncourt, Lee, and Szolovits}{Dernoncourt
  et~al\mbox{.}}{2017}]%
        {dernoncourt2017neuroner}
\bibfield{author}{\bibinfo{person}{Franck Dernoncourt},
  \bibinfo{person}{Ji~Young Lee}, {and} \bibinfo{person}{Peter Szolovits}.}
  \bibinfo{year}{2017}\natexlab{}.
\newblock \showarticletitle{{NeuroNER}: an easy-to-use program for named-entity
  recognition based on neural networks}.
\newblock \bibinfo{journal}{\emph{Conference on Empirical Methods on Natural
  Language Processing (EMNLP)}} (\bibinfo{year}{2017}).
\newblock


\bibitem[\protect\citeauthoryear{D’hondt, Grouin, and Grau}{D’hondt
  et~al\mbox{.}}{2017}]%
        {d2017generating}
\bibfield{author}{\bibinfo{person}{Eva D’hondt}, \bibinfo{person}{Cyril
  Grouin}, {and} \bibinfo{person}{Brigitte Grau}.}
  \bibinfo{year}{2017}\natexlab{}.
\newblock \showarticletitle{Generating a training corpus for OCR
  post-correction using encoder-decoder model}. In
  \bibinfo{booktitle}{\emph{Proceedings of the Eighth International Joint
  Conference on Natural Language Processing (Volume 1: Long Papers)}}.
  \bibinfo{pages}{1006--1014}.
\newblock


\bibitem[\protect\citeauthoryear{{Etter}, {Rawls}, {Carpenter}, and
  {Sell}}{{Etter} et~al\mbox{.}}{2019}]%
        {syntheticrecipeocr}
\bibfield{author}{\bibinfo{person}{D. {Etter}}, \bibinfo{person}{S. {Rawls}},
  \bibinfo{person}{C. {Carpenter}}, {and} \bibinfo{person}{G. {Sell}}.}
  \bibinfo{year}{2019}\natexlab{}.
\newblock \showarticletitle{A Synthetic Recipe for OCR}. In
  \bibinfo{booktitle}{\emph{2019 International Conference on Document Analysis
  and Recognition (ICDAR)}}. \bibinfo{pages}{864--869}.
\newblock


\bibitem[\protect\citeauthoryear{Hakala, Vesanto, Miekka, Salakoski, and
  Ginter}{Hakala et~al\mbox{.}}{2019}]%
        {hakala2019leveraging}
\bibfield{author}{\bibinfo{person}{Kai Hakala}, \bibinfo{person}{Aleksi
  Vesanto}, \bibinfo{person}{Niko Miekka}, \bibinfo{person}{Tapio Salakoski},
  {and} \bibinfo{person}{Filip Ginter}.} \bibinfo{year}{2019}\natexlab{}.
\newblock \showarticletitle{Leveraging Text Repetitions and Denoising
  Autoencoders in OCR Post-correction}.
\newblock \bibinfo{journal}{\emph{arXiv preprint arXiv:1906.10907}}
  (\bibinfo{year}{2019}).
\newblock


\bibitem[\protect\citeauthoryear{Hamdi, Jean-Caurant, Sidere, Coustaty, and
  Doucet}{Hamdi et~al\mbox{.}}{2019}]%
        {hamdi2019analysis}
\bibfield{author}{\bibinfo{person}{Ahmed Hamdi}, \bibinfo{person}{Axel
  Jean-Caurant}, \bibinfo{person}{Nicolas Sidere}, \bibinfo{person}{Micka{\"e}l
  Coustaty}, {and} \bibinfo{person}{Antoine Doucet}.}
  \bibinfo{year}{2019}\natexlab{}.
\newblock \showarticletitle{An Analysis of the Performance of Named Entity
  Recognition over OCRed Documents}. In \bibinfo{booktitle}{\emph{2019 ACM/IEEE
  Joint Conference on Digital Libraries (JCDL)}}. IEEE,
  \bibinfo{pages}{333--334}.
\newblock


\bibitem[\protect\citeauthoryear{Hermann, Kocisky, Grefenstette, Espeholt, Kay,
  Suleyman, and Blunsom}{Hermann et~al\mbox{.}}{2015}]%
        {hermann2015teaching}
\bibfield{author}{\bibinfo{person}{Karl~Moritz Hermann}, \bibinfo{person}{Tomas
  Kocisky}, \bibinfo{person}{Edward Grefenstette}, \bibinfo{person}{Lasse
  Espeholt}, \bibinfo{person}{Will Kay}, \bibinfo{person}{Mustafa Suleyman},
  {and} \bibinfo{person}{Phil Blunsom}.} \bibinfo{year}{2015}\natexlab{}.
\newblock \showarticletitle{Teaching machines to read and comprehend}. In
  \bibinfo{booktitle}{\emph{Advances in neural information processing
  systems}}. \bibinfo{pages}{1693--1701}.
\newblock


\bibitem[\protect\citeauthoryear{Hochreiter and Schmidhuber}{Hochreiter and
  Schmidhuber}{1997}]%
        {hochreiter1997long}
\bibfield{author}{\bibinfo{person}{Sepp Hochreiter} {and}
  \bibinfo{person}{J{\"u}rgen Schmidhuber}.} \bibinfo{year}{1997}\natexlab{}.
\newblock \showarticletitle{Long short-term memory}.
\newblock \bibinfo{journal}{\emph{Neural computation}} \bibinfo{volume}{9},
  \bibinfo{number}{8} (\bibinfo{year}{1997}), \bibinfo{pages}{1735--1780}.
\newblock


\bibitem[\protect\citeauthoryear{Huang, Xu, and Yu}{Huang
  et~al\mbox{.}}{2015}]%
        {huang2015bidirectional}
\bibfield{author}{\bibinfo{person}{Zhiheng Huang}, \bibinfo{person}{Wei Xu},
  {and} \bibinfo{person}{Kai Yu}.} \bibinfo{year}{2015}\natexlab{}.
\newblock \showarticletitle{Bidirectional LSTM-CRF models for sequence
  tagging}.
\newblock \bibinfo{journal}{\emph{arXiv preprint arXiv:1508.01991}}
  (\bibinfo{year}{2015}).
\newblock


\bibitem[\protect\citeauthoryear{Hwang, Kim, Seo, Yim, Park, Park, Lee, Lee,
  and Lee}{Hwang et~al\mbox{.}}{2019}]%
        {hwang2019postocr}
\bibfield{author}{\bibinfo{person}{Wonseok Hwang}, \bibinfo{person}{Seonghyeon
  Kim}, \bibinfo{person}{Minjoon Seo}, \bibinfo{person}{Jinyeong Yim},
  \bibinfo{person}{Seunghyun Park}, \bibinfo{person}{Sungrae Park},
  \bibinfo{person}{Junyeop Lee}, \bibinfo{person}{Bado Lee}, {and}
  \bibinfo{person}{Hwalsuk Lee}.} \bibinfo{year}{2019}\natexlab{}.
\newblock \showarticletitle{Post-{\{}OCR{\}} parsing: building simple and
  robust parser via {\{}BIO{\}} tagging}. In \bibinfo{booktitle}{\emph{Workshop
  on Document Intelligence at NeurIPS 2019}}.
\newblock
\urldef\tempurl%
\url{https://openreview.net/forum?id=SJgjf695UB}
\showURL{%
\tempurl}


\bibitem[\protect\citeauthoryear{Jatowt, Coustaty, Nguyen, Doucet,
  et~al\mbox{.}}{Jatowt et~al\mbox{.}}{2019}]%
        {jatowt2019deep}
\bibfield{author}{\bibinfo{person}{Adam Jatowt}, \bibinfo{person}{Mickael
  Coustaty}, \bibinfo{person}{Nhu-Van Nguyen}, \bibinfo{person}{Antoine
  Doucet}, {et~al\mbox{.}}} \bibinfo{year}{2019}\natexlab{}.
\newblock \showarticletitle{Deep statistical analysis of OCR errors for
  effective post-OCR processing}. In \bibinfo{booktitle}{\emph{2019 ACM/IEEE
  Joint Conference on Digital Libraries (JCDL)}}. IEEE,
  \bibinfo{pages}{29--38}.
\newblock


\bibitem[\protect\citeauthoryear{Jean-Caurant, Tamani, Courboulay, and
  Burie}{Jean-Caurant et~al\mbox{.}}{2017}]%
        {jean2017lexicographical}
\bibfield{author}{\bibinfo{person}{Axel Jean-Caurant},
  \bibinfo{person}{Nouredine Tamani}, \bibinfo{person}{Vincent Courboulay},
  {and} \bibinfo{person}{Jean-Christophe Burie}.}
  \bibinfo{year}{2017}\natexlab{}.
\newblock \showarticletitle{Lexicographical-Based Order for Post-OCR Correction
  of Named Entities}. In \bibinfo{booktitle}{\emph{2017 14th IAPR International
  Conference on Document Analysis and Recognition (ICDAR)}},
  Vol.~\bibinfo{volume}{1}. IEEE, \bibinfo{pages}{1192--1197}.
\newblock


\bibitem[\protect\citeauthoryear{Journet, Visani, Mansencal, Van-Cuong, and
  Billy}{Journet et~al\mbox{.}}{2017}]%
        {journet2017doccreator}
\bibfield{author}{\bibinfo{person}{Nicholas Journet}, \bibinfo{person}{Muriel
  Visani}, \bibinfo{person}{Boris Mansencal}, \bibinfo{person}{Kieu Van-Cuong},
  {and} \bibinfo{person}{Antoine Billy}.} \bibinfo{year}{2017}\natexlab{}.
\newblock \showarticletitle{Doccreator: A new software for creating synthetic
  ground-truthed document images}.
\newblock \bibinfo{journal}{\emph{Journal of imaging}} \bibinfo{volume}{3},
  \bibinfo{number}{4} (\bibinfo{year}{2017}), \bibinfo{pages}{62}.
\newblock


\bibitem[\protect\citeauthoryear{Miller, Boisen, Schwartz, Stone, and
  Weischedel}{Miller et~al\mbox{.}}{2000}]%
        {miller2000named}
\bibfield{author}{\bibinfo{person}{David Miller}, \bibinfo{person}{Sean
  Boisen}, \bibinfo{person}{Richard Schwartz}, \bibinfo{person}{Rebecca Stone},
  {and} \bibinfo{person}{Ralph Weischedel}.} \bibinfo{year}{2000}\natexlab{}.
\newblock \showarticletitle{Named entity extraction from noisy input: speech
  and OCR}. In \bibinfo{booktitle}{\emph{Sixth Applied Natural Language
  Processing Conference}}. \bibinfo{pages}{316--324}.
\newblock


\bibitem[\protect\citeauthoryear{Miller, Vandome, and McBrewster}{Miller
  et~al\mbox{.}}{2009}]%
        {levenshtein}
\bibfield{author}{\bibinfo{person}{Frederic~P. Miller},
  \bibinfo{person}{Agnes~F. Vandome}, {and} \bibinfo{person}{John McBrewster}.}
  \bibinfo{year}{2009}\natexlab{}.
\newblock \bibinfo{booktitle}{\emph{Levenshtein Distance: Information Theory,
  Computer Science, String (Computer Science), String Metric,
  Damerau?Levenshtein Distance, Spell Checker, Hamming Distance}}.
\newblock \bibinfo{publisher}{Alpha Press}.
\newblock
\showISBNx{6130216904}


\bibitem[\protect\citeauthoryear{Needleman and Wunsch}{Needleman and
  Wunsch}{1970}]%
        {needleman1970general}
\bibfield{author}{\bibinfo{person}{SB Needleman} {and} \bibinfo{person}{CD
  Wunsch}.} \bibinfo{year}{1970}\natexlab{}.
\newblock \showarticletitle{A General Method Applicable to the Search for
  Similarities in the Amino Acid Sequence of Two Proteins}.
\newblock \bibinfo{journal}{\emph{Journal of Molecular Biology}}
  (\bibinfo{year}{1970}).
\newblock


\bibitem[\protect\citeauthoryear{Omelianchuk, Atrasevych, Chernodub, and
  Skurzhanskyi}{Omelianchuk et~al\mbox{.}}{2020}]%
        {omelianchuk2020gector}
\bibfield{author}{\bibinfo{person}{Kostiantyn Omelianchuk},
  \bibinfo{person}{Vitaliy Atrasevych}, \bibinfo{person}{Artem Chernodub},
  {and} \bibinfo{person}{Oleksandr Skurzhanskyi}.}
  \bibinfo{year}{2020}\natexlab{}.
\newblock \showarticletitle{GECToR--Grammatical Error Correction: Tag, Not
  Rewrite}.
\newblock \bibinfo{journal}{\emph{arXiv preprint arXiv:2005.12592}}
  (\bibinfo{year}{2020}).
\newblock


\bibitem[\protect\citeauthoryear{Packer, Lutes, Stewart, Embley, Ringger,
  Seppi, and Jensen}{Packer et~al\mbox{.}}{2010}]%
        {packer2010extracting}
\bibfield{author}{\bibinfo{person}{Thomas~L Packer}, \bibinfo{person}{Joshua~F
  Lutes}, \bibinfo{person}{Aaron~P Stewart}, \bibinfo{person}{David~W Embley},
  \bibinfo{person}{Eric~K Ringger}, \bibinfo{person}{Kevin~D Seppi}, {and}
  \bibinfo{person}{Lee~S Jensen}.} \bibinfo{year}{2010}\natexlab{}.
\newblock \showarticletitle{Extracting person names from diverse and noisy OCR
  text}. In \bibinfo{booktitle}{\emph{Proceedings of the fourth workshop on
  Analytics for noisy unstructured text data}}. ACM, \bibinfo{pages}{19--26}.
\newblock


\bibitem[\protect\citeauthoryear{Pradhan, Moschitti, Xue, Uryupina, and
  Zhang}{Pradhan et~al\mbox{.}}{2012}]%
        {pradhan2012conll}
\bibfield{author}{\bibinfo{person}{Sameer Pradhan}, \bibinfo{person}{Alessandro
  Moschitti}, \bibinfo{person}{Nianwen Xue}, \bibinfo{person}{Olga Uryupina},
  {and} \bibinfo{person}{Yuchen Zhang}.} \bibinfo{year}{2012}\natexlab{}.
\newblock \showarticletitle{CoNLL-2012 shared task: Modeling multilingual
  unrestricted coreference in OntoNotes}. In \bibinfo{booktitle}{\emph{Joint
  Conference on EMNLP and CoNLL-Shared Task}}. Association for Computational
  Linguistics, \bibinfo{pages}{1--40}.
\newblock


\bibitem[\protect\citeauthoryear{Sang and De~Meulder}{Sang and
  De~Meulder}{2003}]%
        {sang2003introduction}
\bibfield{author}{\bibinfo{person}{Erik Tjong~Kim Sang} {and}
  \bibinfo{person}{Fien De~Meulder}.} \bibinfo{year}{2003}\natexlab{}.
\newblock \showarticletitle{Introduction to the CoNLL-2003 Shared Task:
  Language-Independent Named Entity Recognition}. In
  \bibinfo{booktitle}{\emph{Proceedings of the Seventh Conference on Natural
  Language Learning at HLT-NAACL 2003}}. \bibinfo{pages}{142--147}.
\newblock


\bibitem[\protect\citeauthoryear{Suissa, Elmalech, and
  Zhitomirsky-Geffet}{Suissa et~al\mbox{.}}{2020}]%
        {suissa2020optimizing}
\bibfield{author}{\bibinfo{person}{Omri Suissa}, \bibinfo{person}{Avshalom
  Elmalech}, {and} \bibinfo{person}{Maayan Zhitomirsky-Geffet}.}
  \bibinfo{year}{2020}\natexlab{}.
\newblock \showarticletitle{Optimizing the neural network training for OCR
  error correction of historical Hebrew texts}.
\newblock \bibinfo{journal}{\emph{iConference 2020 Proceedings}}
  (\bibinfo{year}{2020}).
\newblock


\bibitem[\protect\citeauthoryear{van Strien., Beelen., Ardanuy., Hosseini.,
  McGillivray., and Colavizza.}{van Strien. et~al\mbox{.}}{2020}]%
        {strien2020ocrimpact}
\bibfield{author}{\bibinfo{person}{Daniel van Strien.}, \bibinfo{person}{Kaspar
  Beelen.}, \bibinfo{person}{Mariona~Coll Ardanuy.}, \bibinfo{person}{Kasra
  Hosseini.}, \bibinfo{person}{Barbara McGillivray.}, {and}
  \bibinfo{person}{Giovanni Colavizza.}} \bibinfo{year}{2020}\natexlab{}.
\newblock \showarticletitle{Assessing the Impact of OCR Quality on Downstream
  NLP Tasks}. In \bibinfo{booktitle}{\emph{Proceedings of the 12th
  International Conference on Agents and Artificial Intelligence - Volume 1:
  ARTIDIGH,}}. INSTICC, \bibinfo{publisher}{SciTePress},
  \bibinfo{pages}{484--496}.
\newblock
\showISBNx{978-989-758-395-7}
\urldef\tempurl%
\url{https://doi.org/10.5220/0009169004840496}
\showDOI{\tempurl}


\bibitem[\protect\citeauthoryear{Vaswani, Shazeer, Parmar, Uszkoreit, Jones,
  Gomez, Kaiser, and Polosukhin}{Vaswani et~al\mbox{.}}{2017}]%
        {vaswani2017attention}
\bibfield{author}{\bibinfo{person}{Ashish Vaswani}, \bibinfo{person}{Noam
  Shazeer}, \bibinfo{person}{Niki Parmar}, \bibinfo{person}{Jakob Uszkoreit},
  \bibinfo{person}{Llion Jones}, \bibinfo{person}{Aidan~N Gomez},
  \bibinfo{person}{{\L}ukasz Kaiser}, {and} \bibinfo{person}{Illia
  Polosukhin}.} \bibinfo{year}{2017}\natexlab{}.
\newblock \showarticletitle{Attention is all you need}. In
  \bibinfo{booktitle}{\emph{Advances in neural information processing
  systems}}. \bibinfo{pages}{5998--6008}.
\newblock


\bibitem[\protect\citeauthoryear{Wang, Zhang, Ren, Zhang, Zitnik, Shang,
  Langlotz, and Han}{Wang et~al\mbox{.}}{2019}]%
        {wang2019cross}
\bibfield{author}{\bibinfo{person}{Xuan Wang}, \bibinfo{person}{Yu Zhang},
  \bibinfo{person}{Xiang Ren}, \bibinfo{person}{Yuhao Zhang},
  \bibinfo{person}{Marinka Zitnik}, \bibinfo{person}{Jingbo Shang},
  \bibinfo{person}{Curtis Langlotz}, {and} \bibinfo{person}{Jiawei Han}.}
  \bibinfo{year}{2019}\natexlab{}.
\newblock \showarticletitle{Cross-type biomedical named entity recognition with
  deep multi-task learning}.
\newblock \bibinfo{journal}{\emph{Bioinformatics}} \bibinfo{volume}{35},
  \bibinfo{number}{10} (\bibinfo{year}{2019}), \bibinfo{pages}{1745--1752}.
\newblock


\bibitem[\protect\citeauthoryear{Yalniz and Manmatha}{Yalniz and
  Manmatha}{2011}]%
        {retas}
\bibfield{author}{\bibinfo{person}{Ismet~Zeki Yalniz} {and}
  \bibinfo{person}{Raghavan Manmatha}.} \bibinfo{year}{2011}\natexlab{}.
\newblock \showarticletitle{A fast alignment scheme for automatic ocr
  evaluation of books}. In \bibinfo{booktitle}{\emph{2011 International
  Conference on Document Analysis and Recognition}}. IEEE,
  \bibinfo{pages}{754--758}.
\newblock


\end{thebibliography}

\appendix

\section{Document Generation}
\label{sec:document-generation}
\begin{figure*}[ht]
    \centering
    \subcaptionbox{Multi-column Document Template\label{fig:multi_column_tempalte}}[0.3\linewidth]{
        \includegraphics[clip, trim=0cm 13cm 0cm 0cm, width=\linewidth]{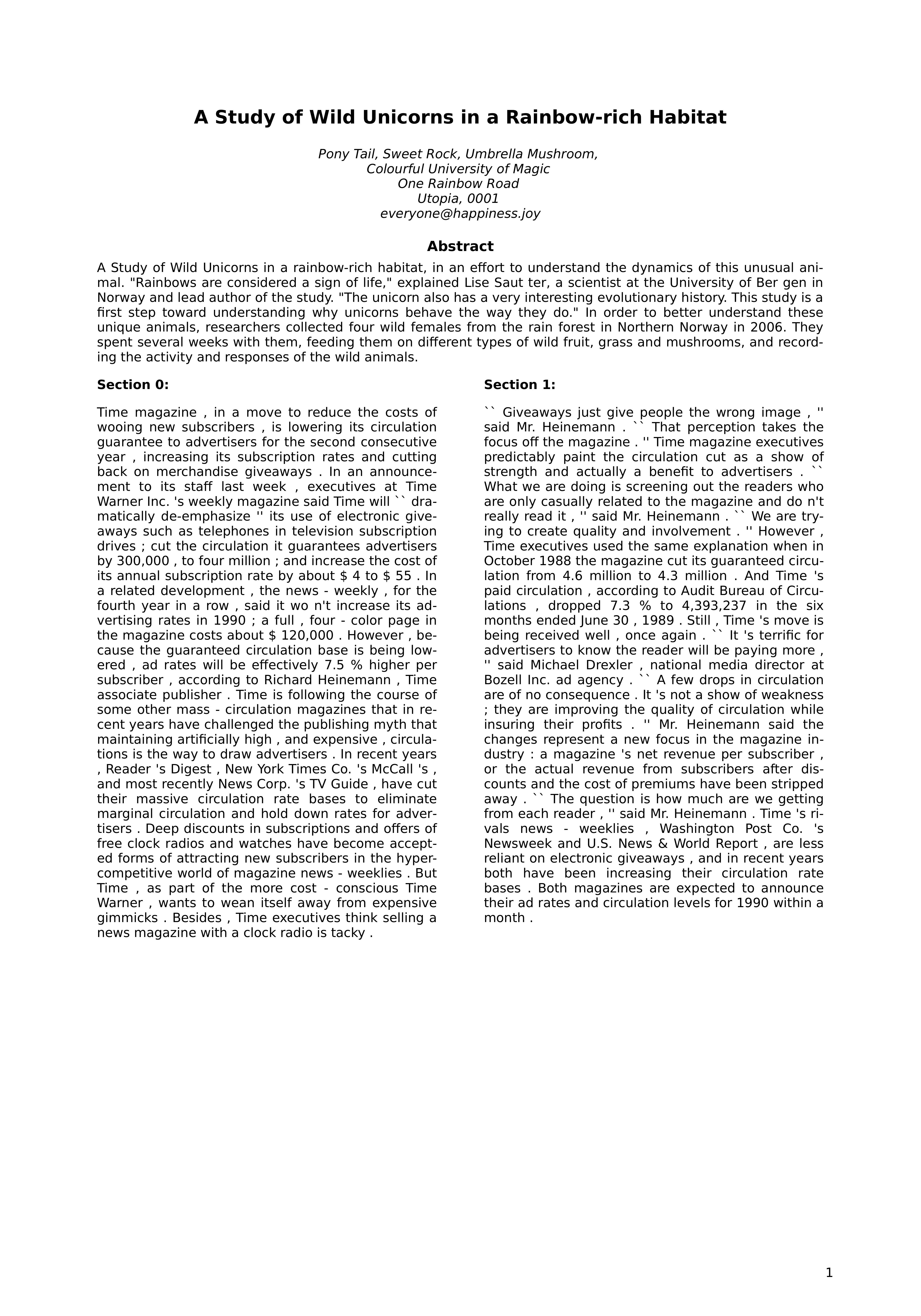}
    }
    \hfill
    \subcaptionbox{Letter-like Document Template\label{fig:letter_like_tempalte}}[0.3\linewidth]{
        \includegraphics[clip, trim=0cm 13cm 0cm 0cm, width=\linewidth]{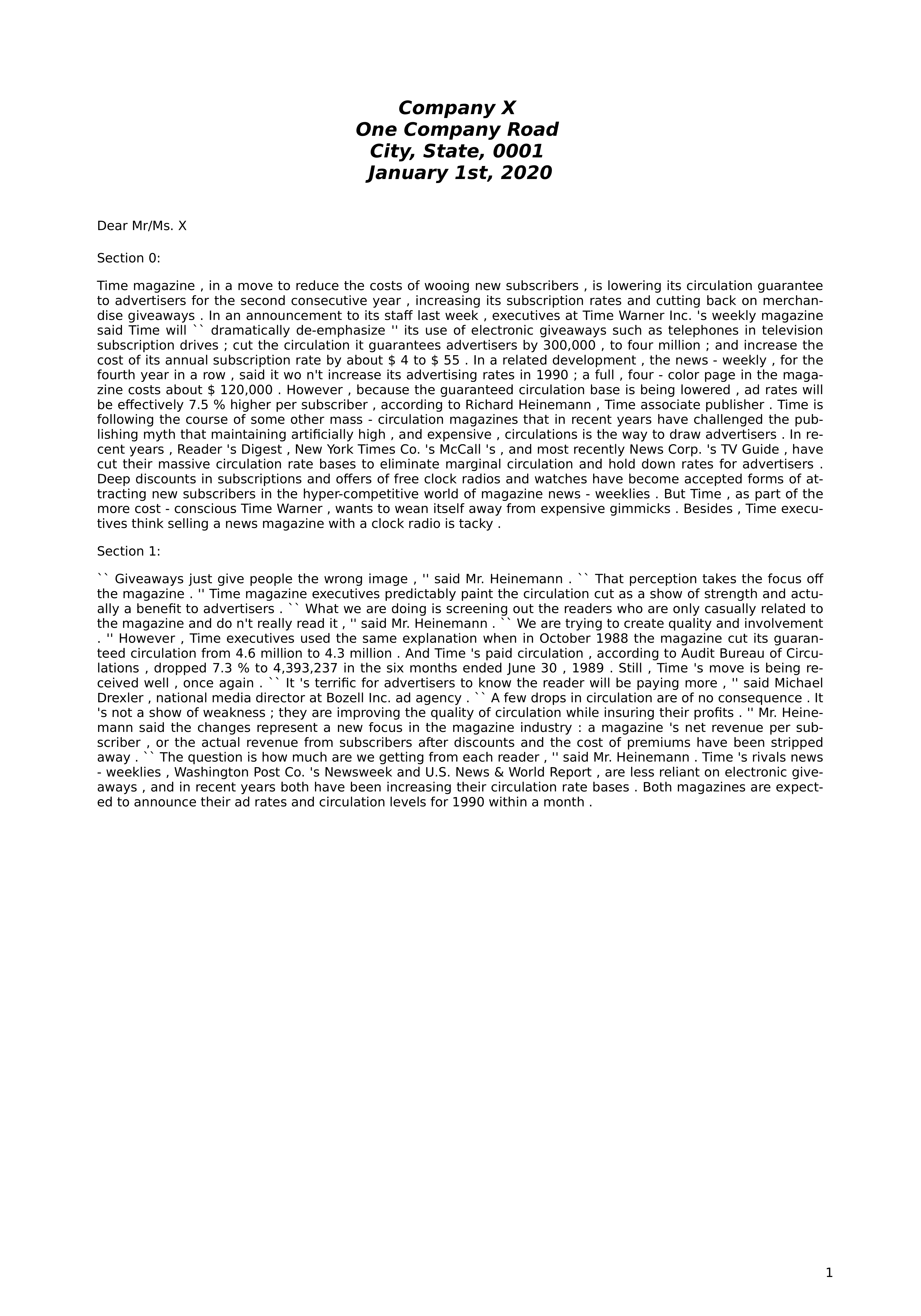}
    }
    \hfill
    \subcaptionbox{Simple Text Block Template\label{fig:text_block_template}}[0.3\linewidth]{
        \includegraphics[clip, trim=0cm 13cm 0cm 0cm, width=\linewidth]{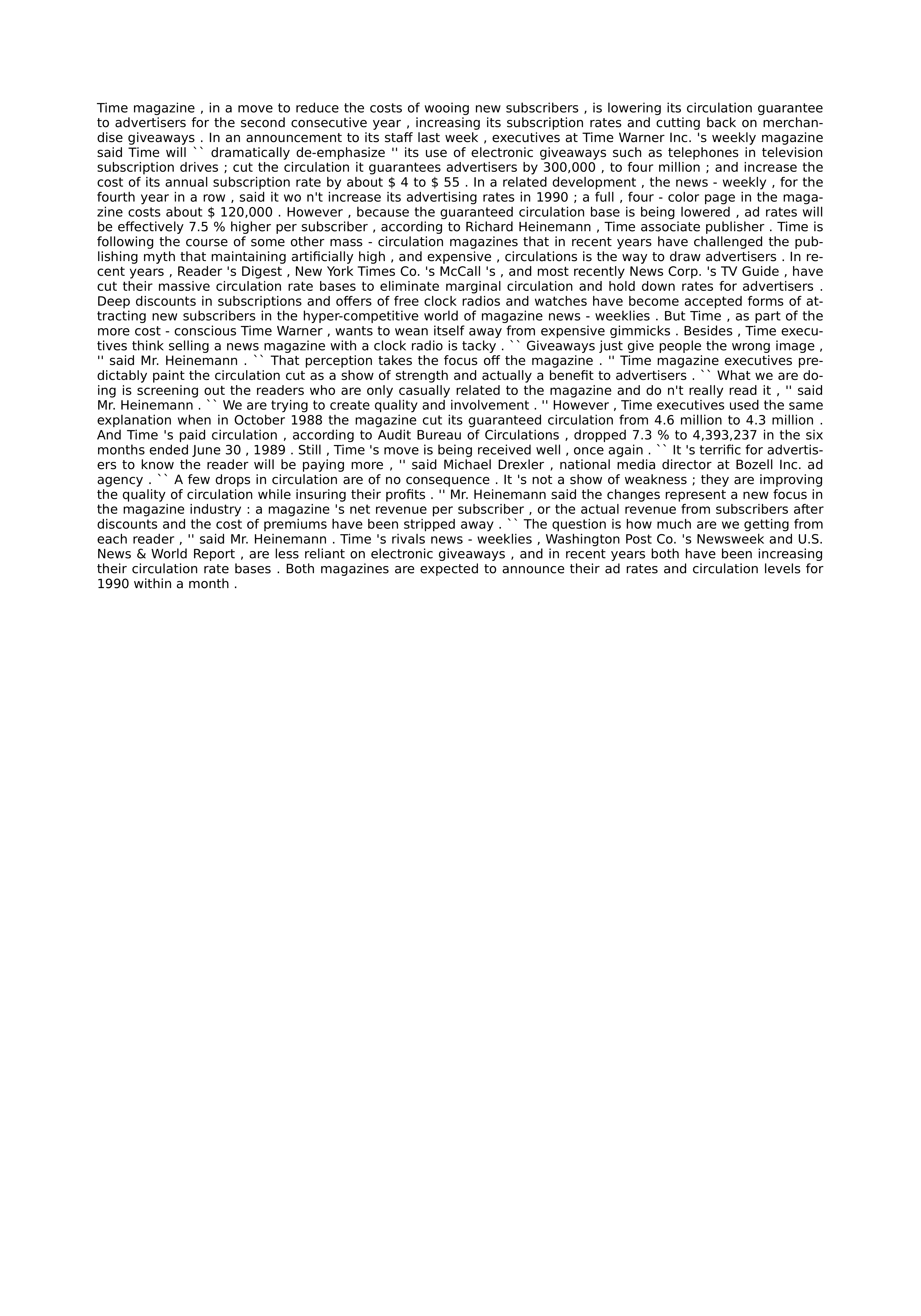}
    }
    \caption{The three standard document templates available in \texttt{Genalog}}\label{fig:templates}
\end{figure*}

\texttt{Genalog} provides three standard document templates for document generation, depicted on~\autoref{fig:templates}.

\section{Document Degradation}
\label{sec:image-degradation}

\begin{figure*}[htp!]
    \centering
    \subcaptionbox{No Degradation\label{fig:no-degradations}\vspace{0.4cm}}[0.3\linewidth]{
        \includegraphics[clip, trim=0cm 0cm 0cm 0cm, width=\linewidth]{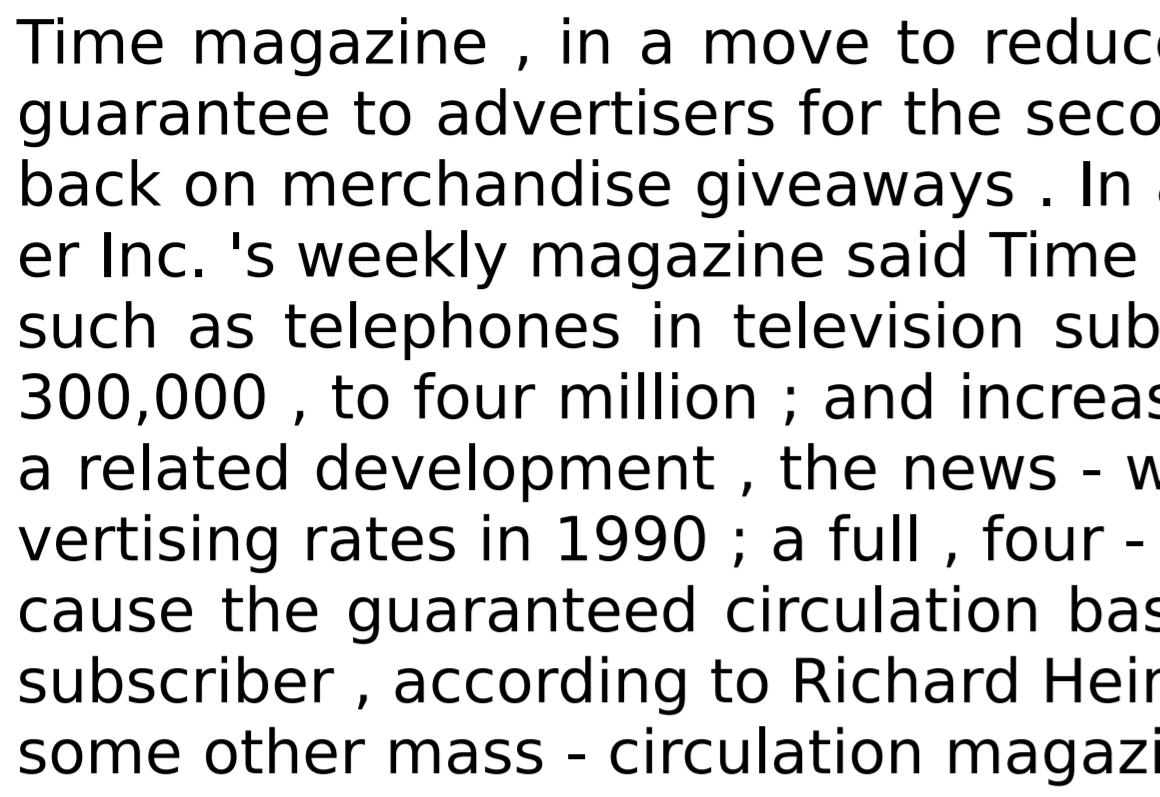}
    }
    \hfill
    \subcaptionbox{All Degradations (Light) \label{fig:all-degradations-light}\vspace{0.4cm}}[0.3\linewidth]{
        \includegraphics[clip, trim=0cm 0cm 0cm 0cm, width=\linewidth]{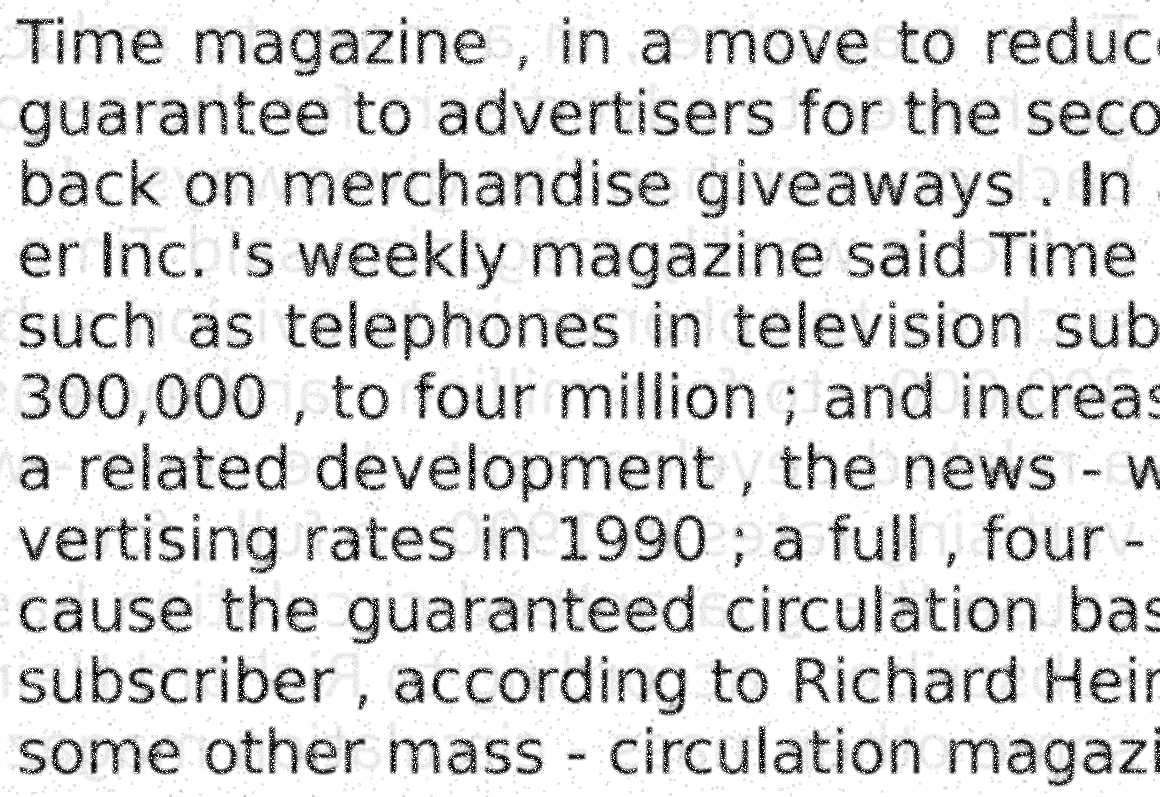}
    }    
    \hfill
    \subcaptionbox{All Degradations (Heavy)\label{fig:all-degradations-heavy}\vspace{0.4cm}}[0.3\linewidth]{
        \includegraphics[clip, trim=0cm 0cm 0cm 0cm, width=\linewidth]{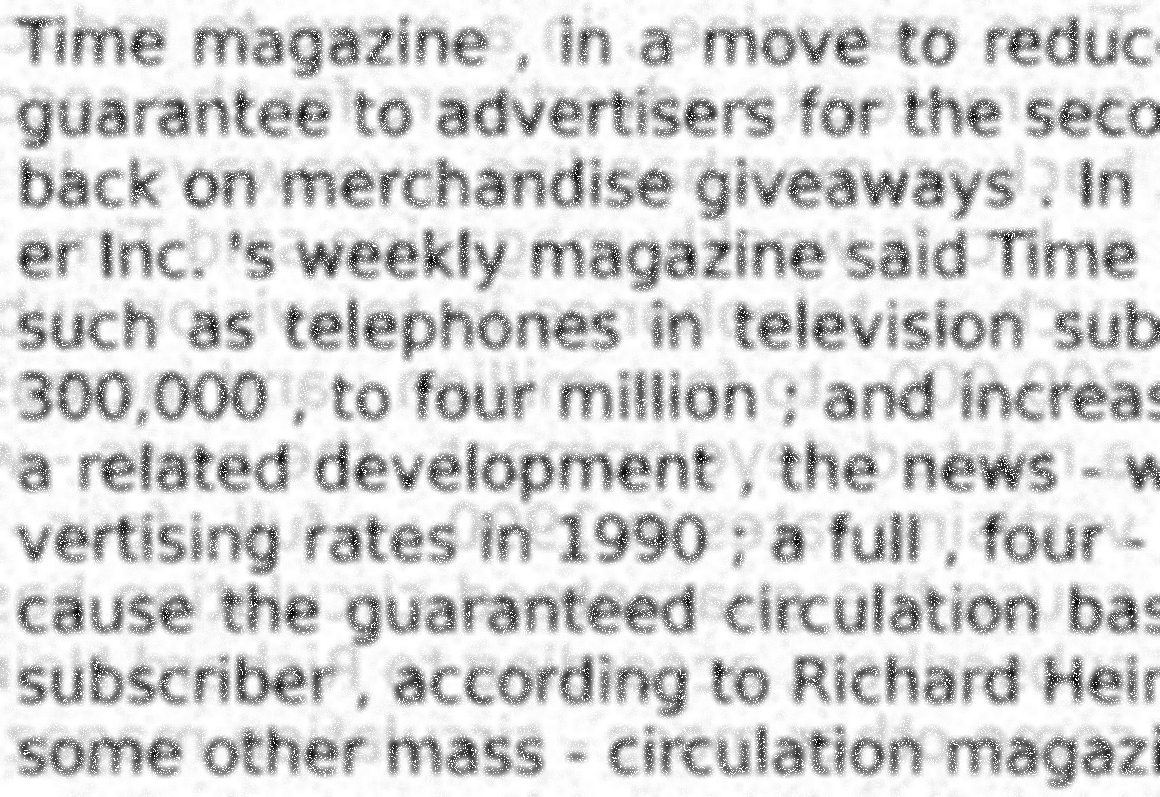}
    }
    \subcaptionbox{Bleed-through (Heavy)\label{fig:degradations_bleed_heavy}\vspace{0.4cm}}[0.3\linewidth]{
        \includegraphics[clip, trim=0cm 0cm 0cm 0cm, width=\linewidth]{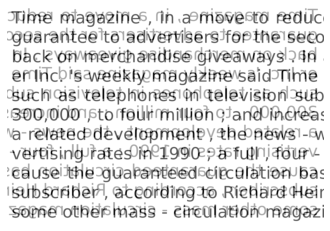}
    }
    \hfill
    \subcaptionbox{Blur (Heavy)\label{fig:degradations_blur_heavy}\vspace{0.4cm}}[0.3\linewidth]{
        \includegraphics[clip, trim=0cm 0cm 0cm 0cm, width=\linewidth]{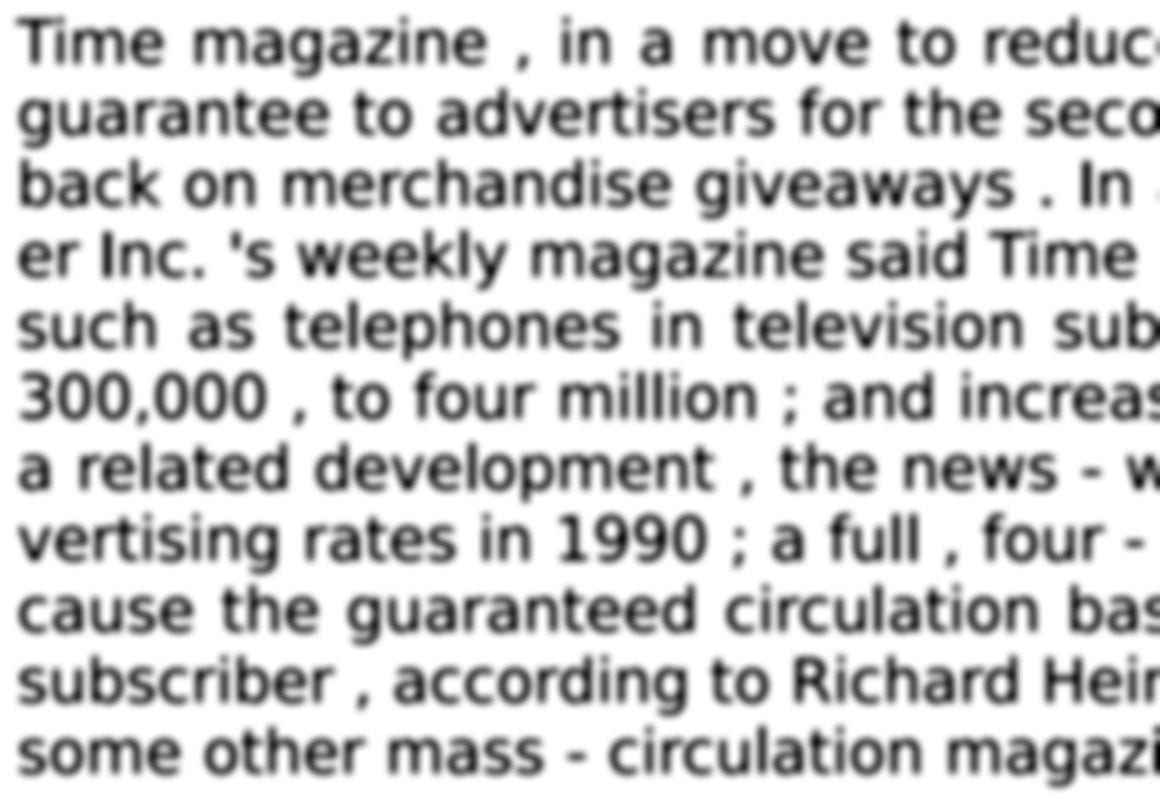}
    }
    \hfill
    \subcaptionbox{Open (Heavy)\label{fig:degradations_open_heavy}\vspace{0.4cm}}[0.3\linewidth]{
        \includegraphics[clip, trim=0cm 0cm 0cm 0cm, width=\linewidth]{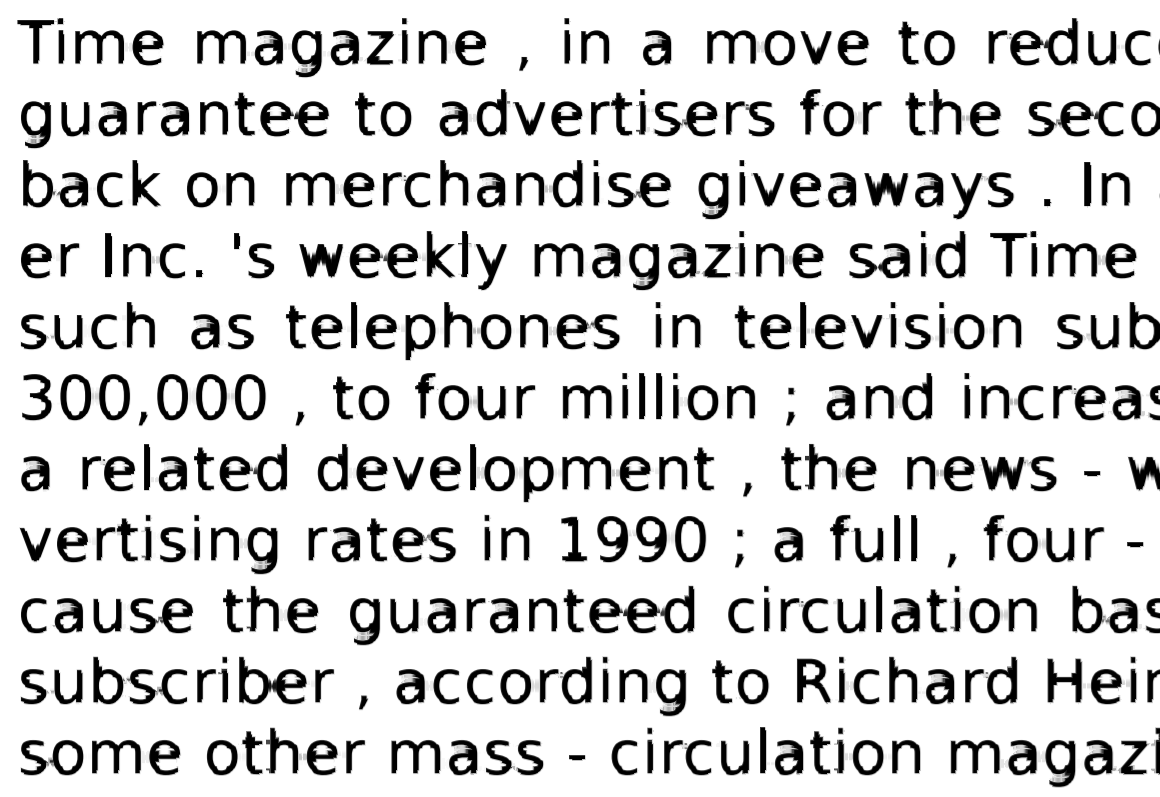}
    }
    \subcaptionbox{Bleed-through (Light)\label{fig:degradations_bleed_light}\vspace{0.4cm}}[0.3\linewidth]{
        \includegraphics[clip, trim=0cm 0cm 0cm 0cm, width=\linewidth]{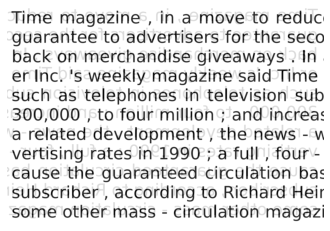}
    }
    \hfill
    \subcaptionbox{Blur (Light)\label{fig:degradations_blur_light}\vspace{0.4cm}}[0.3\linewidth]{
        \includegraphics[clip, trim=0cm 0cm 0cm 0cm, width=\linewidth]{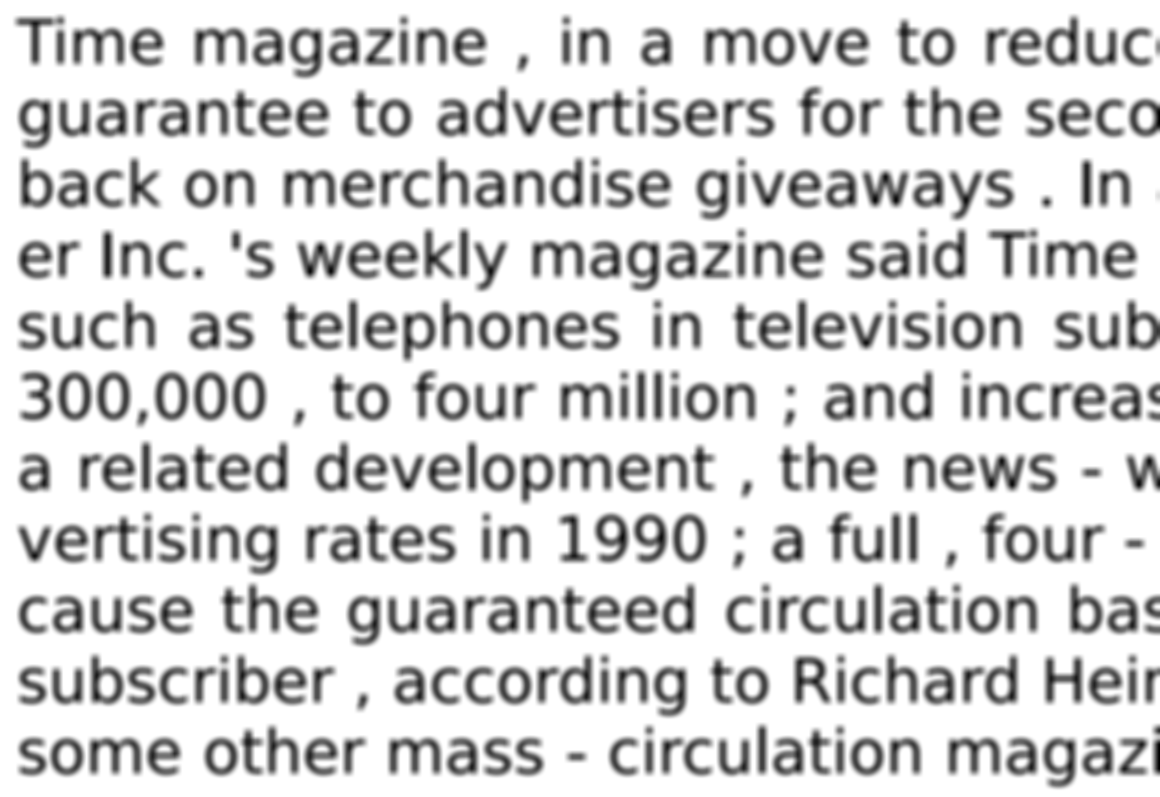}
    }
    \hfill
    \subcaptionbox{Open (Light)\label{fig:degradations_open_light}\vspace{0.4cm}}[0.3\linewidth]{
        \includegraphics[clip, trim=0cm 0cm 0cm 0cm, width=\linewidth]{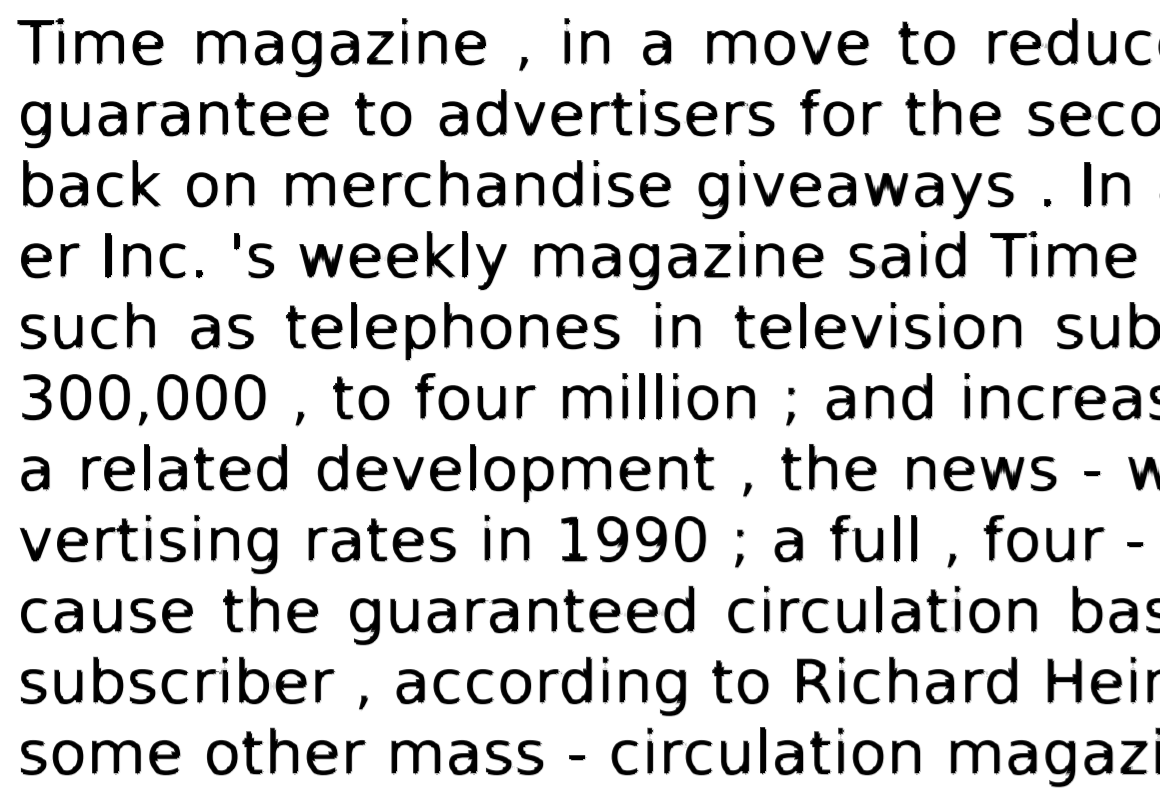}
    }
    \hfill
    \subcaptionbox{Close (Heavy)\label{fig:degradations_close_heavy}\vspace{0.4cm}}[0.3\linewidth]{
        \includegraphics[clip, trim=0cm 0cm 0cm 0cm, width=\linewidth]{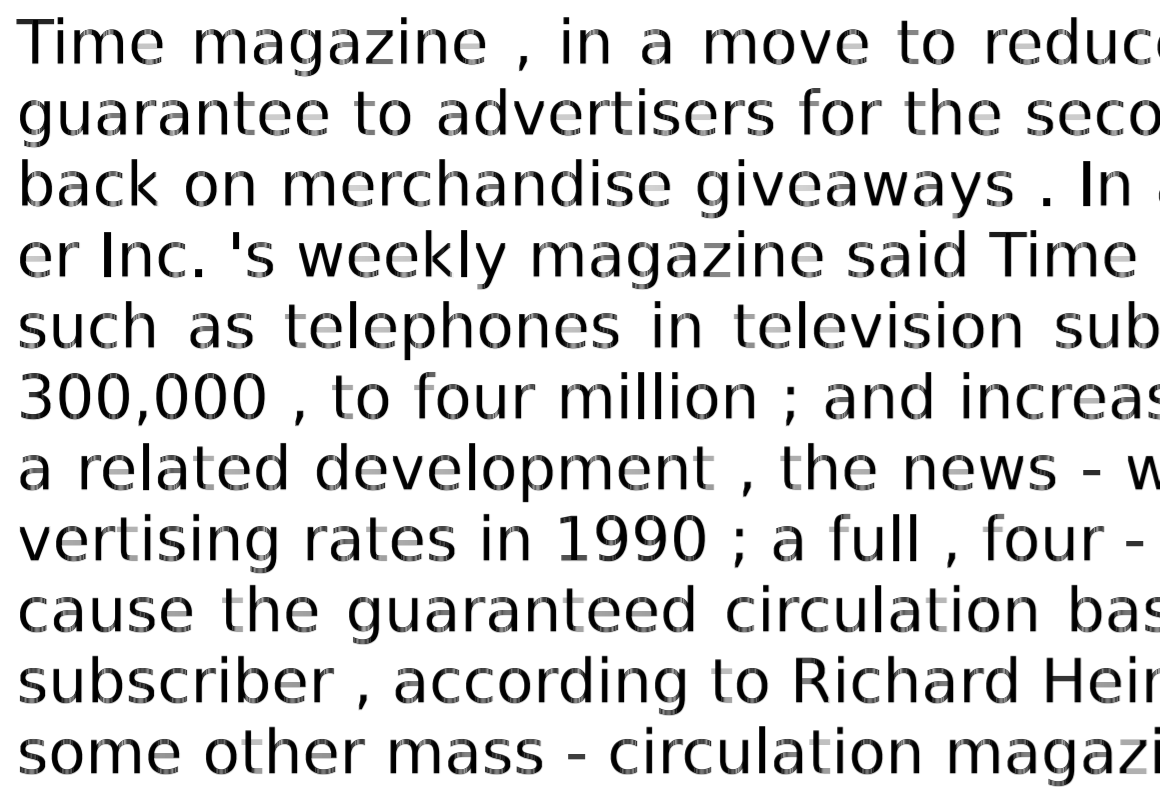}
    }
    \hfill
    \subcaptionbox{Pepper (Heavy)\label{fig:degradations_pepper_heavy}\vspace{0.4cm}}[0.3\linewidth]{
        \includegraphics[clip, trim=0cm 0cm 0cm 0cm, width=\linewidth]{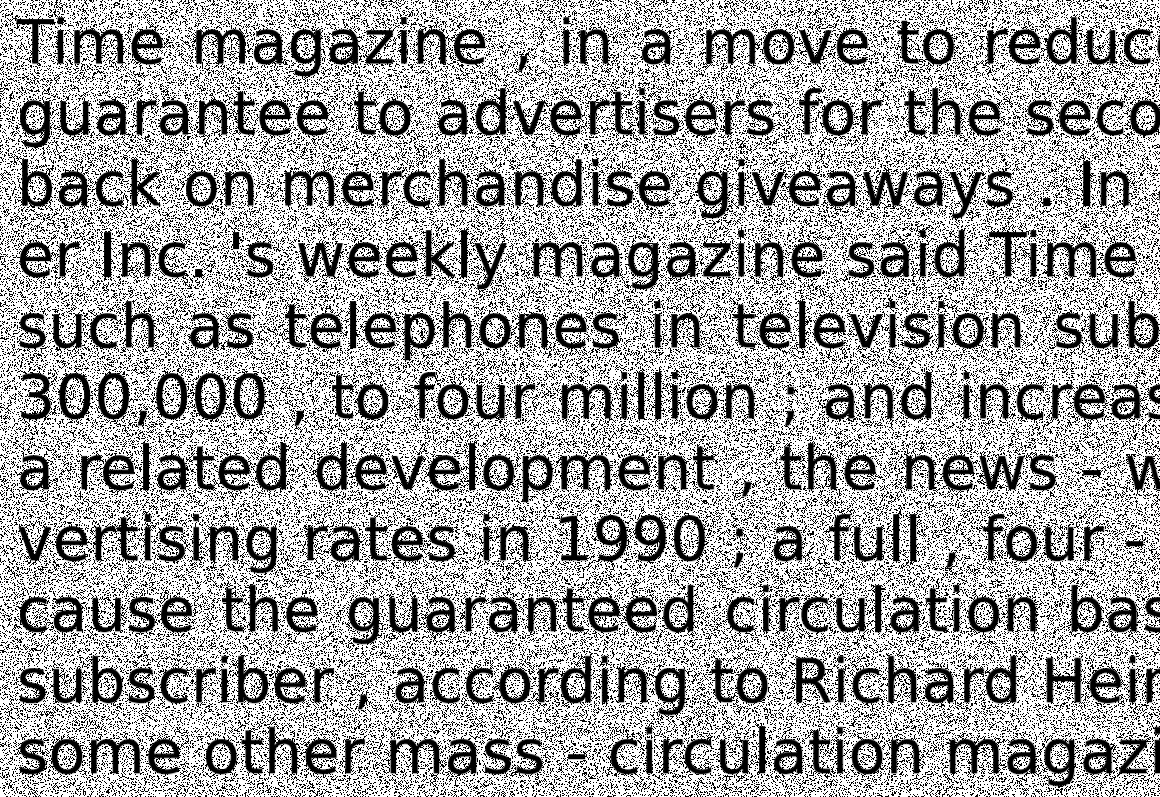}
    }
    \hfill
    \subcaptionbox{Salt (Heavy)\label{fig:degradations_salt_heavy}\vspace{0.4cm}}[0.3\linewidth]{
        \includegraphics[clip, trim=0cm 0cm 0cm 0cm, width=\linewidth]{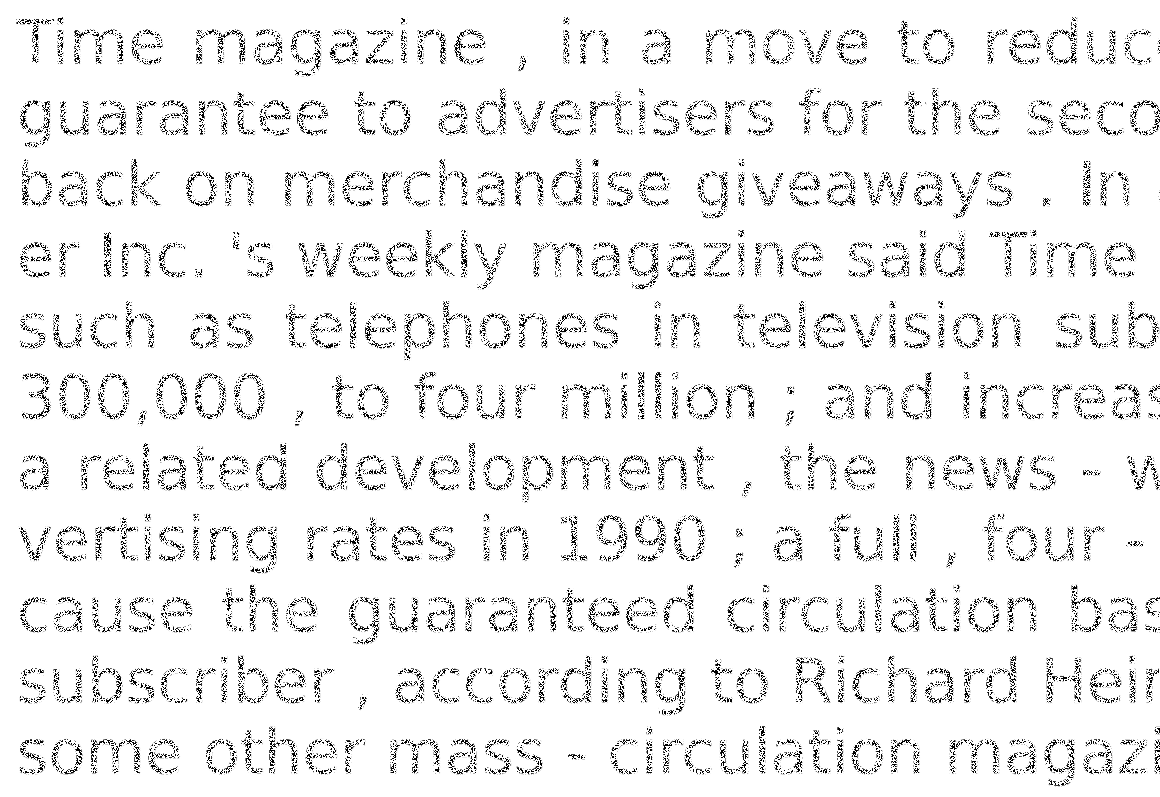}
    }
    \hfill
    \subcaptionbox{Close (Light)\label{fig:degradations_close_light}\vspace{0.4cm}}[0.3\linewidth]{
        \includegraphics[clip, trim=0cm 0cm 0cm 0cm, width=\linewidth]{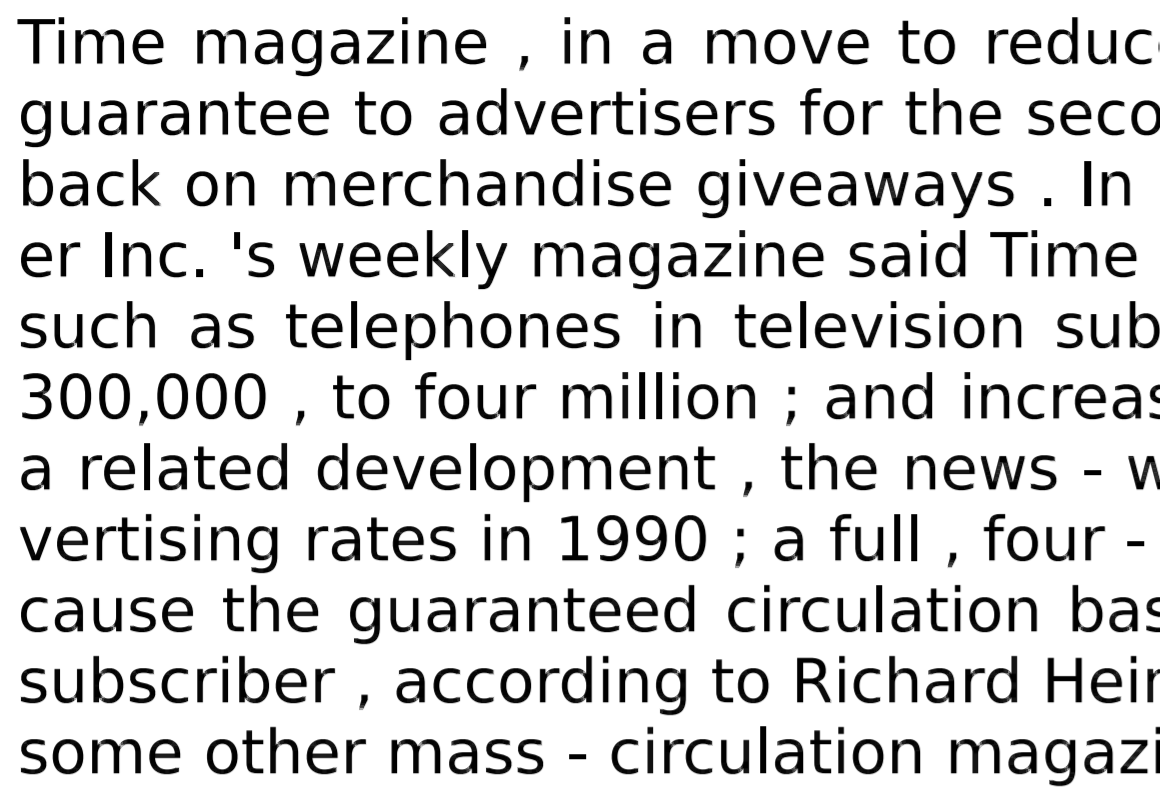}
    }
    \hfill
    \subcaptionbox{Pepper (Light)\label{fig:degradations_pepper_light}\vspace{0.4cm}}[0.3\linewidth]{
        \includegraphics[clip, trim=0cm 0cm 0cm 0cm, width=\linewidth]{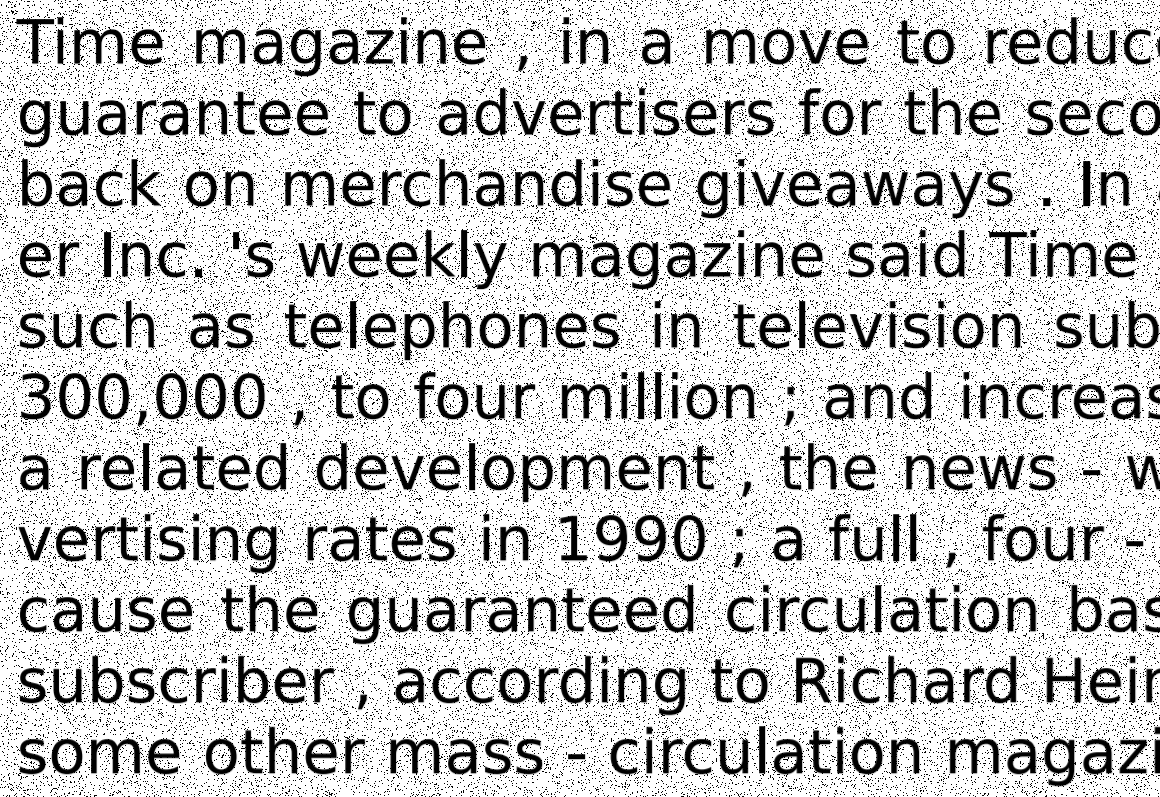}
    }
    \hfill
    \subcaptionbox{Salt (Light)\label{fig:degradations_salt_light}\vspace{0.4cm}}[0.3\linewidth]{
        \includegraphics[clip, trim=0cm 0cm 0cm 0cm, width=\linewidth]{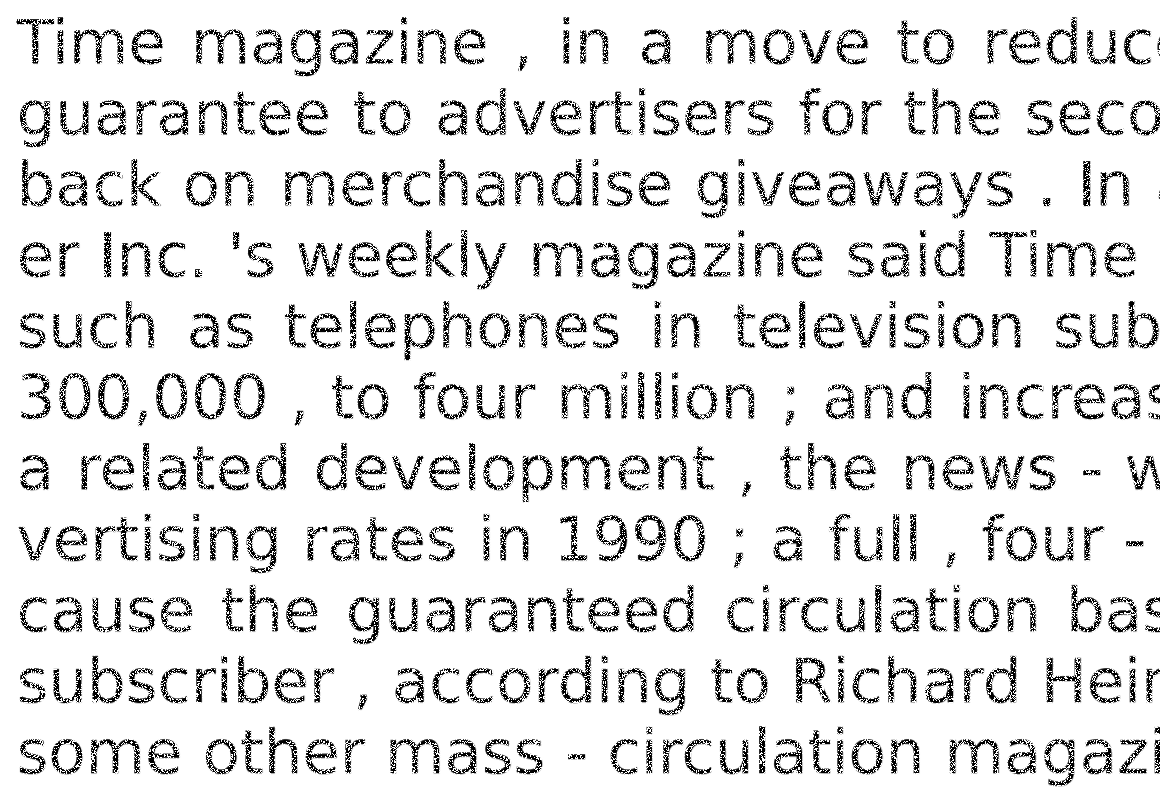}
    }
    \caption{Example of degradations produced by \texttt{Genalog}}
    \label{fig:appendix-degradation-examples}
\end{figure*}

We show example degradations produced by \texttt{Genalog} in~\autoref{fig:appendix-degradation-examples}.

\begin{table*}[htp]
\centering
\resizebox{\textwidth}{!}{%

\begin{tabular}{@{}lllllllll@{}}
\toprule
Degradation             & No Degradation & Bleed Heavy & Blur Heavy & Close Heavy & Open Heavy & Pepper Heavy  & Salt Heavy & All Heavy \\ \midrule
Word Accuracy & 0.954          & 0.735       & 0.902       & 0.831       & 0.940      & 0.879        & 0.905         & 0.697         \\
Character Accuracy & 0.991          & 0.945       & 0.980      & 0.966       & 0.988      & 0.975        & 0.984         & 0.915           \\\bottomrule
\end{tabular}%

}
\caption{OCR accuracy on degraded text, obtained from the CoNLL-2012 train set. Examples of the degraded documents are shown in~\autoref{fig:appendix-degradation-examples}.}
\label{tbl:results_ocr_horizontal}
\end{table*}

\section{Degradation Parameters of Used Datasets}
\label{sec:appendix-degradation-params}

\autoref{tab:all-degradation-params} lists the degradation parameters used in our experiments.

\begin{table*}[ht]
\begin{tabular}{@{}lllll@{}}
\toprule
Order & Degradation   & Parameter Name & Parameter Value & Description                              \\ \midrule
1     & open          & kernel shape   & (9,9)           & Thickens characters            \\
      & -             & kernel type    & plus            & Kernel filled with "1"s in a "+" shape   \\
2     & close         & kernel shape   & (9,1)           & Remove horizontal structures \\
      & -             & kernel type    & ones            & Kernel filled all with "1"s              \\
3     & salt          & amount         & 0.5 / 0.7             & Substantially reduce effect with more salt  \\
4     & overlay       & src            & ORIGINAL\textunderscore STATE & Overlay current image on the original    \\
      & -             & background     & CURRENT\textunderscore STATE  &                                          \\
5     & bleed through & src            & CURRENT\textunderscore STATE  & Apply bleed-through effect               \\
      & -             & background     & ORIGINAL\textunderscore STATE &                                          \\
      & -             & alpha          & 0.8 / 0.8             & Transparency factor                         \\
      & -             & offset x       & -6 / -5              & Offset (pixels) of the background in x-axis \\
      & -             & offset y       & -12 / -5             & Offset (pixels) of the background in y-axis \\
6     & pepper        & amount         & 0.015 / 0.005           & Imitate page degradation                 \\
7     & blur          & radius         & 11 / 3              & Apply Gaussian blur                      \\
8     & salt          & amount         & 0.15            & Add digital noise        \\ \bottomrule               
\end{tabular}
\caption{The sequence, in order, of types of degradation, along with the associated parameter values that was utilized to produce the All Degradations (Heavy / Light) dataset. Please see~\autoref{fig:all-degradations-heavy} and~\autoref{fig:all-degradations-light} for example images. Note ORIGINAL\_STATE refers to the original state of an image before any degradation, and CURRENT\_STATE refers to the current state of an image after the last degradation operation.}
\label{tab:all-degradation-params}
\end{table*}

\end{document}